\documentclass[11pt]{article}
\usepackage[margin=1in]{geometry}
\usepackage{amsmath,amssymb,amsthm,booktabs,url}
\usepackage{graphicx}
\usepackage[colorlinks=true,linkcolor=blue!50!black,citecolor=blue!50!black,
            urlcolor=blue!50!black]{hyperref}
\usepackage{xcolor}
\usepackage{array}
\usepackage{tikz}
\usetikzlibrary{arrows.meta}
\newtheorem{lemma}{Lemma}
\newtheorem{definition}{Definition}

\title{Runtime Observability for Heterogeneous Attention Memory}
\author{%
  Fanzhe Wei\\ Metask Lab\\ {\small\texttt{whyer1@gmail.com}}
  \and Li Liu\\ Metask Lab\\ {\small\texttt{muriel092611@gmail.com}}
  \and Ziyang Wang\\
  {\small\begin{tabular}[t]{@{}c@{}}
   Logistics Science and Technology Innovation Integrated\\
   Development Research Center of Shaanxi Logistics Group,\\
   Xi'an Jiaotong University, Xi'an, China\\
   and\\
   School of Management,\\
   Xi'an Jiaotong University, Xi'an, China\\[2pt]
   \texttt{wang-zy13@tsinghua.org.cn}
   \end{tabular}}
  \and Chenyu Wang\\ Metask Lab\\ {\small\texttt{cnados@gmail.com}}}
\date{}

\begin{document}
\maketitle

\begin{abstract}
Modern models no longer keep a plain KV cache: latent caches, learned sparse
selectors and recurrent states each carry the model's memory in a different
form, and each fails differently under compression. We give a runtime
observability contract for all four memory classes, built from three
operators (\textsc{update}/\textsc{select}/\textsc{read}) and instantiated
on six model configurations across five architecture families. Three
results. \emph{Typed composition}: contracts carry their error metric as a
type, so a cross-metric chain is rejected at construction---this check
refuted our own first composed chain---and every claim carries a
machine-decided tier (certified, partially certified, or empirical), with
composition inheriting the weakest tier. \emph{An executable risk ledger}:
replayed over $12.4$M entry reads and run under eight-way concurrency with
per-request budgets and fail-closed identity attribution, the ledger holds
its risk budget with zero violations, and a fused always-on probe observes
a declared one-layer subset under CUDA graphs inside the serving noise
floor. \emph{Diagnosis on a served stack}: applied to DeepSeek-V4 with a
packed compressed-KV prototype, the same machinery localizes a silent
corruption to a precise structural boundary---exact in the eviction-free,
identity-isolated regime, with every observed failure under eviction or
slot reuse---through a machine-adjudicated discrimination campaign whose
calculus rejected two of our own confounded inferences along the way. All
artifacts, guards, and the Lean development are released at
\url{https://github.com/metask-ai/witprobe-attention-memory}; every number
in this paper regenerates from the shipped artifacts by one command.
\end{abstract}

\section{Introduction}
\label{sec:intro}
On 2026-06-25 a silent corruption in a hierarchical cache took down agentic traffic
on one of our production hosts. Nothing in the serving stack was in a position to
notice: the cache was structurally valid, the model kept generating, and the failure
surfaced only as degraded task success far downstream. That incident is the
motivation for this paper. A compressed or shared attention memory is state that the
system trusts without verifying, and the trust is currently blind.

The obvious response---``check the KV cache''---no longer names a single object.
Contemporary architectures keep their attention memory in at least four different
forms: dense per-token key/value tensors; a latent cache decompressed on read; a
learned selector that picks a sparse subset of entries; and a recurrent state
carried forward rather than stored. They fail differently, they admit different
guarantees, and a bound proved for one says nothing about another.

We make four contributions.

\textbf{A contract for attention memory.} We write any attention memory as three
operators---update, select, read---and give each stage a local contract
$e_{\mathrm{out}}\le a\,e_{\mathrm{in}}+b$ with an explicit failure budget $\delta$
(\S\ref{sec:contract}). Supporting a new architecture becomes a proof obligation
rather than a new bespoke tool.

\textbf{Composition with a machine-decidable strength.} Local contracts compose by
affine composition and a union bound. The contribution is not the algebra, which is
elementary, but that the platform \emph{computes} the composition from live probe
readings and labels every number certified, partially certified or empirical.
Composition inherits the weakest tier, so ``end-to-end certified'' is a statement the
system refuses to print unless every stage earns it. This caught a live error in our
own chain: an early version used the \emph{true} residual as the storage bound, a
quantity that requires an uncompressed reference and therefore cannot be computed at
runtime at all.

\textbf{Cross-architecture evidence with an always-on cost.} We instantiate the
platform on six models across five architecture families
(\S\ref{sec:platform}, \S\ref{sec:eval}), report per-stage readings with their
measurement regimes, and measure the cost of leaving the probes on under concurrency.

\textbf{Diagnosis on a served stack.} Pointed at a served DeepSeek-V4 stack with a
packed compressed-KV prototype, the same machinery ran a certificate-guided format
exploration, localized a silent corruption to a precise structural boundary through
a machine-adjudicated discrimination campaign---whose calculus rejected two of our
own confounded inferences---and served the packed path end to end under a six-path
serving-tax taxonomy (\S\ref{sec:v4}).

\textbf{Artifacts.} The measurement mathematics, the acceptance gates,
every frozen run artifact behind every number in this paper, the figure
and claim generators with their guards, and the complete Lean
development (including the experiment-adjudication calculus) are
released as \texttt{witprobe-attention-memory}
(\url{https://github.com/metask-ai/witprobe-attention-memory}) under
Apache-2.0; every
number regenerates from the shipped artifacts by one command, and the
claim guard fails the build on any mismatch. The architecture-specific
probe-injection implementation is not part of the release
(\S\ref{sec:limits}).

\textbf{Terminology.} A \emph{contract} is a per-stage guarantee template whose
error metric is part of its type; instantiated on a model, it is monitored
through a \emph{witness}---the locally computable quantity the probes read.
Every emitted number carries a \emph{tier}: \emph{certified} (sound bound, not
saturated), \emph{partially certified} (bound on a decidable subset), or
\emph{empirical} (observation, no bound); composition inherits the weakest tier
(\S\ref{sec:tiers}). The \emph{ledger} is the executable request-level account
that composes per-stage failure budgets into one risk verdict
(\S\ref{sec:ledger}). A \emph{measurement regime} is the serving condition under
which a number was obtained---serial or concurrent, CUDA graphs on or off,
eviction-free or under eviction; numbers are comparable only within a regime,
and every reading below is annotated with its regime.

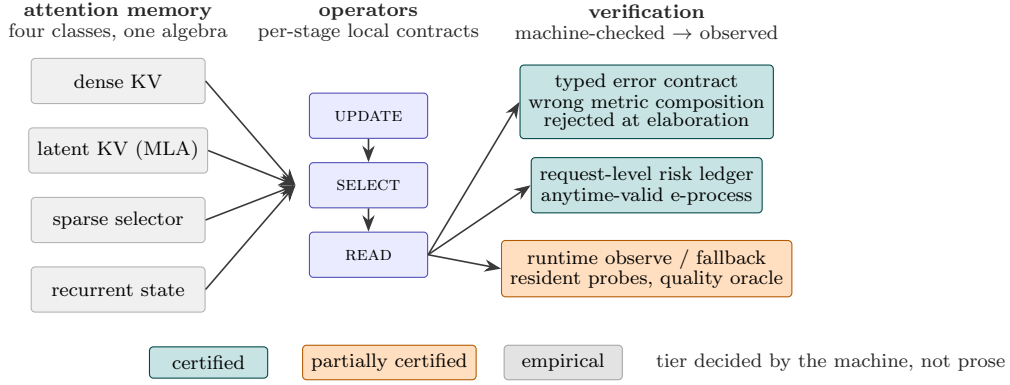
\begin{figure}[t]
\centering
\begin{tikzpicture}[font=\scriptsize, scale=0.92, every node/.append style={transform shape},
  box/.style={draw, rounded corners=1.5pt, align=center, inner sep=3.5pt,
              minimum height=6.5mm},
  mem/.style={box, fill=gray!12, draw=gray!60, minimum width=25mm},
  op/.style={box, fill=blue!8, draw=blue!40!gray, minimum width=17mm},
  cert/.style={box, fill=teal!22, draw=teal!55!black},
  part/.style={box, fill=orange!25, draw=orange!70!black},
  emp/.style={box, fill=gray!22, draw=gray!70},
  arr/.style={-{Stealth[length=2.2mm]}, semithick, gray!45!black}]

\node[mem] (dense)  at (0, 3.0) {dense KV};
\node[mem] (latent) at (0, 2.0) {latent KV (MLA)};
\node[mem] (sparse) at (0, 1.0) {sparse selector};
\node[mem] (rec)    at (0, 0.0) {recurrent state};
\node[align=center, gray!30!black] at (0, 3.85) {\textbf{attention memory}\\[-1pt]\scriptsize four classes, one algebra};

\node[op] (upd) at (3.6, 2.5) {\textsc{update}};
\node[op] (sel) at (3.6, 1.5) {\textsc{select}};
\node[op] (rd)  at (3.6, 0.5) {\textsc{read}};
\node[align=center, gray!30!black] at (3.6, 3.85) {\textbf{operators}\\[-1pt]\scriptsize per-stage local contracts};

\foreach \m in {dense, latent, sparse, rec}{
  \draw[arr] (\m.east) -- (2.55, 1.5);}
\draw[arr] (upd) -- (sel);
\draw[arr] (sel) -- (rd);

\node[cert, minimum width=31mm] (typed)  at (7.6, 2.7)
  {typed error contract\\[-1pt]\scriptsize wrong metric composition\\[-2pt]\scriptsize rejected at elaboration};
\node[cert, minimum width=31mm] (ledger) at (7.6, 1.5)
  {request-level risk ledger\\[-1pt]\scriptsize anytime-valid e-process};
\node[part, minimum width=31mm] (obs)    at (7.6, 0.3)
  {runtime observe / fallback\\[-1pt]\scriptsize resident probes, quality oracle};
\draw[arr] (rd.east) -- (typed.west);
\draw[arr] (rd.east) -- (ledger.west);
\draw[arr] (rd.east) -- (obs.west);
\node[align=center, gray!30!black] at (7.6, 3.85) {\textbf{verification}\\[-1pt]\scriptsize machine-checked $\to$ observed};

\node[cert, minimum width=17mm, minimum height=4.5mm] (l1) at (1.3, -1.05) {\scriptsize certified};
\node[part, minimum width=22mm, minimum height=4.5mm] (l2) at (3.9, -1.05) {\scriptsize partially certified};
\node[emp,  minimum width=17mm, minimum height=4.5mm] (l3) at (6.4, -1.05) {\scriptsize empirical};
\node[gray!30!black, anchor=west] at (7.6, -1.05) {\scriptsize tier decided by the machine, not prose};
\end{tikzpicture}
\caption{\textbf{WitCert in one view.} Four attention-memory classes, three
operators, typed per-stage contracts, a request-level risk ledger, and
fail-closed runtime observation. Colors mark certification tiers, decided
by machine-checked artifacts (\S\ref{sec:tiers}), not narrative.}
\label{fig:arch}
\end{figure}

\section{Attention Memory Contracts}
\label{sec:contract}

\subsection{Three operators}
Write the memory at step $t$ as $m_t$, and let
\begin{align}
m_{t+1} &= U_t(m_t, x_t) && \text{update / write}\\
I_t     &= S_t(q_t, m_t)  && \text{select (possibly the identity)}\\
y_t     &= R_t(q_t, m_t, I_t) && \text{read}
\end{align}
Compression or corruption replaces $m_t$ by $\hat m_t$; the goal is to bound
$\lVert y_t-\hat y_t\rVert$ from locally computable witnesses. The four memory
objects are instances: a dense KV cache appends and selects everything; a latent
cache appends a compressed representation and reconstructs on read; a sparse
selector adds a non-trivial $S$; a recurrent state makes $U$ a contraction rather
than an append.

\subsection{Local contracts and composition}
\begin{definition}[Contract]
A stage exposes $(a,b,\delta)$ with $e_{\mathrm{out}}\le a\,e_{\mathrm{in}}+b$
holding except with probability $\delta$.
\end{definition}
Series composition and the request-level budget are
\begin{equation}
(a_2,b_2)\circ(a_1,b_1)=(a_2a_1,\ a_2b_1+b_2),
\qquad
\textstyle\sum_i\delta_i\le\delta_{\mathrm{req}} .
\label{eq:compose}
\end{equation}
Both are elementary. What is not elementary in practice is that each $a$, $b$ and
$\delta$ must be \emph{computable from quantities the serving stack actually has},
and that the resulting number must carry an honest strength label.

\subsection{Three tiers, decided by the machine}
\label{sec:tiers}
\begin{itemize}
\item \textbf{certified} --- a sound upper bound exists and is not saturated;
\item \textbf{partially certified} --- the bound holds only on a decidable subset,
      e.g.\ rank-prefix stability holds only for rows whose margin exceeds
      $2\varepsilon$;
\item \textbf{empirical} --- an observation with no bound attached.
\end{itemize}
Every metric the platform emits is registered against one tier together with the
reason it earns that tier, and an unregistered metric is a hard error rather than a
silent demotion. Composition takes the weakest tier of its stages. This is what
makes the distinction operational rather than rhetorical: a chain containing one
empirical stage cannot be reported as certified, whatever the accompanying prose
says.

\section{Local Contracts}
\label{sec:local}

\subsection{Pooling is an attenuator, not an error source}
Block-pooled architectures form one stored entry from $B$ consecutive tokens with
gate weights $w$, $\sum_j w_j=1$, $w\ge 0$. If the tokens carry errors $\Delta_j$
then $\Delta c=\sum_j w_j\Delta_j$ and:

\begin{lemma}[Pooling, two forms]\label{lem:pool}
With $e_{\mathrm{in}}:=(\sum_j\lVert\Delta_j\rVert^2)^{1/2}$,
$\lVert\Delta c\rVert\le\lVert w\rVert_2\,e_{\mathrm{in}}$ by Cauchy--Schwarz, so
$a=\lVert w\rVert_2\le 1$. With $e_{\mathrm{in}}:=\max_j\lVert\Delta_j\rVert$,
convexity gives $a=1$ with no assumption at all.
\end{lemma}

Both forms must be reported: the $\ell_2$ form is tighter but presumes an $\ell_2$
aggregation of the incoming error, and quoting it alone suggests the attenuation is
unconditional. Note that pooling is part of $U$ and acts on \emph{uncompressed}
inputs, so it is not an error source; it governs how token-level errors upstream of
it propagate. Placing it inside the baseline chain, as we initially did, yields a
bound that is both looser and semantically wrong.

\subsection{Selection is additive, not amplifying}
\begin{lemma}[Rank-prefix selection]\label{lem:sel}
Regard the attention distribution as supported on the full key set and zero outside
the selected set. With $I$ the original selection and $\hat I$ the perturbed one,
\begin{equation}
\mathrm{TV}(p,\hat p)\ \le\
\tfrac12\Big(\underbrace{\textstyle\sum_{i\in I\setminus\hat I}p_i}_{m_{\mathrm{out}}}
+\underbrace{\textstyle\sum_{i\in\hat I\setminus I}\hat p_i}_{m_{\mathrm{in}}}\Big)
+\tfrac12\textstyle\sum_{i\in I\cap\hat I}\lvert p_i-\hat p_i\rvert ,
\label{eq:s2}
\end{equation}
hence $(a_S,b_S)=\big(1,\ \tfrac12(m_{\mathrm{out}}+m_{\mathrm{in}})\big)$.
\end{lemma}
The selection stage therefore \emph{adds} a bias equal to half the swapped attention
mass and does not amplify upstream error. This reframes a negative result: asking
whether the entire top-$k$ set is preserved is the wrong question, because strict set
certification is almost never attainable (\S\ref{sec:eval}), whereas the swapped mass
is small and directly measurable.

\subsection{A sampling sentinel with a detection-latency guarantee}
A sentinel that draws $r$ uniform slots \emph{with replacement} per round (matching
the implementation's \texttt{randint}) from $M$ slots of which $B$ are corrupted
misses everything in one round, and over $n$ independent rounds, with
probability
\begin{equation}
q \;=\; \Bigl(1-\frac{B}{M}\Bigr)^{r},
\qquad
\Pr[\text{missed after } n \text{ rounds}] \;=\; q^{\,n}.
\label{eq:sentinelmiss}
\end{equation} (Sampling without replacement would give
the smaller hypergeometric $q$; modelling the implementation one actually runs is
the point.)

\begin{lemma}[Detection latency]\label{lem:sentinel}
For target confidence $1-\delta$, detection of at least one corrupted slot is
guaranteed within $n(\delta)=\lceil\log\delta/\log q\rceil$ rounds.
\end{lemma}
This converts an anecdotal demo into a monitoring guarantee with a tunable budget,
and it explains an observation that would otherwise look like a capability limit
(\S\ref{sec:eval}).

\section{Platform}
\label{sec:platform}
The implementation separates four concerns: a registry declaring where to inject for
each architecture; adapters that extract the tensors and build the validity mask;
meters holding the mathematics, which are architecture-independent; and a runtime
that keys every accumulator by (owner, layer, path) and always emits a coverage
record. Injection is anchored on exact upstream source blocks with an asserted hit
count, and reverts surgically so that two adapters may share a file.

Acceptance is a fixed order---coverage, then magnitude, then soundness---and a
failure at any step invalidates the numbers that follow. This is not a style
preference. All three checks correspond to failure modes we hit in which the
experiment terminated normally and produced plausible output: a probe whose
per-layer state overwrote itself, a probe that read uninitialised device memory
beyond a kernel's valid range, and a probe attached to a write path the model never
takes. Each produced a clean exit and a wrong number.

The platform currently registers six adapters over eleven injection points, covering
the four memory objects.

\section{Evaluation}
\label{sec:eval}

\subsection{Setup}
All serving experiments run on a single node with 8$\times$H200 (143\,GB) GPUs on
sglang 0.5.13.post1. The large-model experiments serve DeepSeek-V4-Flash-FP8 at
tp=8/ep=8 with a 65{,}536-token context; overhead and graph-safety experiments use
Qwen2.5-7B-Instruct on one GPU, where a serving stack restarts in minutes rather
than tens of minutes. Regression runs for smaller models use a separate RTX~4090
machine (sglang 0.5.9). Probes are injected into unmodified serving code by a
line-anchored source injector with hit-count assertions, byte-identical restore
verified against SHA-256 baselines, and a fixed acceptance order---coverage, then
magnitude sanity, then soundness self-checks---that must pass before any number is
read. Unless stated otherwise probes sample one call in $8$ with a 256-row cap per
call, and probe measurements that need host-side I/O run with CUDA graphs disabled;
the graph-enabled configuration is evaluated separately in
\S\ref{sec:overhead}. Every reported number is frozen from a probe artifact by a
canon generator, and a claim guard fails the build if the manuscript and the
artifacts disagree; the guard checks numeric strings, not semantics, so measurement regimes
travel with the numbers in the artifacts themselves.

\subsection{Coverage}
\begin{table}[t]\centering\small
\begin{tabular}{llllr}
\toprule
Model & Family & Probe class & Layers & Key reading\\
\midrule
Qwen2.5-7B      & GQA & KV & 28/28 & residual $0.80\%$, conserv.\ $3.25\times$\\
Llama-3.1-8B    & GQA & KV & 32/32 & residual $1.02\%$, conserv.\ $3.77\times$\\
DeepSeek-V2-Lite& MLA & LatentKV & 27/27 & residual $0.877\%$, conserv.\ $2.69\times$\\
GLM-5.2-W4AFP8  & MLA+DSA & LatentKV & 78/78 & residual $0.551\%$, conserv.\ $2.95\times$\\
GLM-5.2-W4AFP8  & MLA+DSA & SparseSelector & 21/21 & top-1 certified $99.26\%$\\
V4-Flash-FP8    & MLA+entry & LatentKV & 43/43 & residual $0.628\%$, conserv.\ $3.51\times$\\
V4-Flash-FP8    & MLA+entry & SparseSelector & 21/21 & top-1 certified $99.40\%$\\
Kimi-Linear-48B & KDA+MLA & RecurrentState & 20/20 & $\bar a\in[0.530,0.936]$\\
\bottomrule
\end{tabular}
\caption{\textbf{Coverage.} Eight probe rows over six models and
five architecture families, all past the acceptance gate (coverage,
magnitude, soundness); failing rows are not reported.}
\label{tab:coverage}
\end{table}

Table~\ref{tab:coverage} summarises the coverage: \textbf{8 rows / 6 models}
across five families; two models were onboarded without writing any
model-specific code (DeepSeek-V2-Lite needed only the first step of the
onboarding recipe).

\subsection{Per-stage readings}
\textbf{Latent memory.} On the small proxy the latent meter is validated against
per-cell ground truth over $4{,}032$ cells with $0/4{,}032$ Tier-A violations, and
the dithered Tier-B coverage is $98.9\%$ at $\tau=0.2$. In production, GLM-5.2's own
fp8 storage residual is $0.93\%$ of the latent norm, and our secondary int8
quantisation of the same latents measures $0.551\%$ over all $78$ layers. Latent
norms are strongly heterogeneous across depth---a factor of $493\times$ between the
largest and smallest---which is the empirical basis for treating layers
non-uniformly.

\textbf{Sparse selection.} Strict certification of the entire selected set is
essentially unattainable: $0.05\%$ on GLM's token-level top-$2048$. The rank-prefix
view is the opposite: top-1 stability is certified for $99.26\%$ of rows on GLM and
$99.40\%$ on V4's page-level top-$512$, with zero flips inside the certified region.
The two are not directly comparable---one selects tokens and the other pages---so
only the qualitative shape transfers.

\textbf{Pooling.} V4-Flash uses two pooling granularities in the same model,
block $4$ and block $128$, over $21$ and $20$ layers respectively; since
$1/\text{block}$ differs by $32\times$, raw concentration statistics cannot be
compared across them and we report normalised quantities. The normalised
concentration at block $4$ is $0.1551$, i.e.\ close to uniform.

\textbf{Recurrent state.} We report observation only. The per-layer mean contraction
is $\bar a\in[0.530,0.936]$, all below one, but the mean is the wrong statistic:
long-run behaviour is governed by $\sum_t\log a_t$, and by Jensen's inequality
\begin{equation}
\mathbb{E}[\log a] \;\le\; \log\bar a.
\label{eq:jensen}
\end{equation} Measured, $\mathbb{E}[\log a]$ has median
$-0.5912$ against $\log\bar a=-0.2282$, so the mean form \emph{understates} the
contraction by $2.6\times$; the corresponding error half-life is $1.17$ steps rather
than $3.04$. A residual gap remains: between $0.0$ and $28.7\%$ of samples per layer
have $a\ge0.999$, and our aggregate does not retain channel identity, so an aggregate
$\mathbb{E}[\log a]<0$ does not imply per-channel contraction. We therefore do not
claim a certificate here.

\subsection{Composition: the type checker rejected our own chain}
Our first composed number added the storage stage's bound to the selection stage's
$b_S$ by \eqref{eq:compose} and reported the result. That composition is illegal,
and an earlier version of this paper contained it. The storage bound is a
\emph{relative witness} on cache entries---a dimensionless ratio---while $b_S$ is
total variation on an attention distribution, a probability mass; $a_2 b_1 + b_2$
presumes $b_1$ is already denominated in the output metric of stage two, and here it
is not. We caught this by making the metric part of the contract's \emph{type}:
each contract declares an input and an output metric, composition is defined only
when they coincide, and the same rule is enforced twice---at runtime, where the
checker rejects the chain construction, and in Lean, where the ill-typed composition
fails elaboration and the failure itself is pinned as a build assertion. Neither the
tier system nor the numeric claim guard had caught it; both operate on values, and
the values were individually correct. The retracted numbers appear nowhere else in
this paper, and the guard now forbids them.

The repaired chain crosses between metrics through two proved bridges. The score
bridge is Cauchy--Schwarz: a perturbation $\Delta k$ of a cache entry moves the
attention score by at most
\begin{equation}
\bigl|\Delta(\mathrm{scale}\cdot\langle q,k\rangle)\bigr|
\;\le\; \mathrm{scale}\cdot\lVert q\rVert\cdot\lVert\Delta k\rVert.
\label{eq:scorebridge}
\end{equation} Its coefficient is a
runtime quantity, so the probe now collects it: across $43$ layers and $8$ ranks of
V4-Flash, $\mathrm{scale}\cdot\max\lVert q\rVert = 1.0034$---queries pass through
RMSNorm, so the bridge barely amplifies, and the bound is dominated by the witness
$W$ itself (per-layer max up to $0.700$). The distribution bridge instantiates the
machine-checked e-form theorem of the companion certificate paper~\cite{witcert-companion}
(there proved for per-token bounds $c_t$, here specialized to a uniform $\varepsilon$): a score perturbation of at most $\varepsilon$ moves
the softmax distribution by at most
\begin{equation}
\mathrm{TV} \;\le\; \tfrac12\bigl(e^{2\varepsilon}-1\bigr)
\label{eq:tvbridge}
\end{equation}
in total variation. Both directions of enforcement carry the theorem name in the contract's
proof field.

Review of the repaired chain caught a second mislabel, and we keep it on the
record because it is the failure mode the type system cannot see: a hand-written
metric annotation that is simply wrong. Our selection bound $b_S$ is computed from
the softmax of \emph{indexer} logits, so its unit is selector-score mass, not
attention mass; we had typed it as attention-TV by fiat. Retyping it correctly makes
the checker refuse to add $b_S$ to the storage term---the missing piece is an
\emph{empirical} stability bridge from selector scores to realised attention mass,
which no amount of formalisation can conjure. The two are therefore reported
separately and never summed.

What remains in the chain is the window-storage term, and its pairing is exact: the
fused kernel that produces $q$ also stores the window keys $q$ attends to. It gives
a per-layer attention-TV \emph{bound} of $0.786$ under the all-rank
worst-case form (sampled-max coefficients; $29/43$ layers below the trivial $1.0$)
and a \emph{diagnostic reading} of $0.647$ (median) under mean coefficients---the
latter has no bound semantics, since a mean coefficient is not an upper bound, and
we no longer call it one---for the comparison in which window entries are perturbed
and selected pages are treated as exact---the quantisation of the
compressed pages themselves is not yet witnessed, and the softmax bridge needs the
$\ell_\infty$ over \emph{all} logits, so we state that scope rather than claim past
it. Summed over all $43$ layers the budget is $36.3$---vacuous, and
reported as such; the sum is an aggregate over distinct per-layer distributions, not
an end-to-end output distance, and no bridge across depth is claimed.

Both scope restrictions were then discharged by measurement rather than
by more formalism: with every term witnessed on the $21$ sparse-selection
layers the full chain gives a worst-case per-layer TV bound of $1.473$
(median), and the missing selector-to-attention bridge is replaced by a
paired measurement---the ratio of attention mass to selector-score mass is
$0.87$ (median; converted with the tail value $2.18$)---so the selection
term crosses into attention units at tier \emph{empirical}, and the
composed tier drops with it, automatically. The full accounting, including
why these numbers are \emph{larger} than the retracted ones (they pay the
full price of soundness), is Appendix~\ref{app:composition}.

\subsection{An executable request-level ledger}\label{sec:ledger}
\begin{figure}[t]\centering
\includegraphics[width=0.52\linewidth]{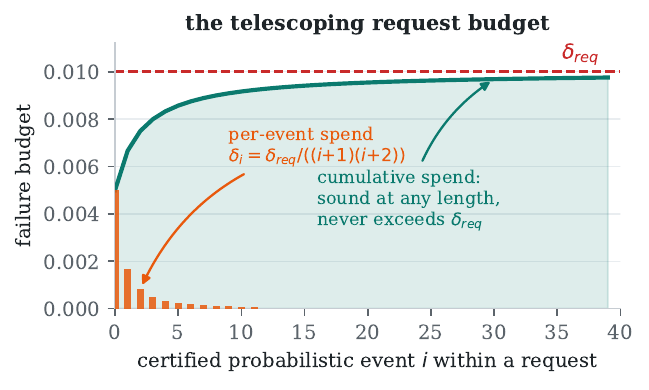}
\caption{\textbf{The telescoping request budget.} Per-event spends
$\delta_i=\delta_{req}/((i{+}1)(i{+}2))$ (orange) and their cumulative
sum (teal) never reach $\delta_{req}$ (dashed): sound at unknown request
length---the analytic content of \texttt{telescope\_sum}, drawn at the
implementation's $\delta_{req}=0.01$.}
\label{fig:budget}
\end{figure}

The per-stage contracts become operational through a request-level risk ledger with
three rules: empirical objects never enter the certified ledger (they either fall
back or degrade the request's verdict); deterministic certificates---runtime
checks whose premises the machine can decide---enter at $\delta=0$; and
probabilistic certificates draw from a telescoping budget
\begin{equation}
\delta_i \;=\; \frac{\delta_{\mathrm{req}}}{(i{+}1)(i{+}2)},
\qquad \sum_{i\ge 0}\delta_i \;<\; \delta_{\mathrm{req}},
\label{eq:telescope}
\end{equation}
so the guarantee holds at every request length without knowing the
length in advance. The budget rules are machine-checked
theorems (\texttt{ledger\_sound}, \texttt{telescope\_sum},
\texttt{two\_layer\_budget\_le}; Figure~\ref{fig:budget} shows the
telescoping spend); the exclusion rule is not a theorem but an API
discipline, enforced by the ledger implementation and locked by regression tests. The guarantee this machinery is built to deliver is the first tier of
request-level claims: every read that uses a compressed result carries a valid
local certificate, with failure probability at most $\delta_{\mathrm{req}}$. What
this paper demonstrates is deliberately less, in two stages. An offline replay
estimates the coverage of a prospective per-entry certified fallback policy over
probe-sampled reads. An \emph{online shadow ledger} then closes four of the six
gaps between replay and execution: ledgers are instantiated per request with real
request boundaries, every selected-page entry read at every decode step of every
sparse-selection layer is checked online ($198{,}163{,}408$ entry events over
$24$ serial requests), and each request ends with its own verdict---median
per-request retention $0.334$ at the same threshold, with a hypothetical
page-in of $5.4$\,GiB per request pricing what real fallback would move.

Certificate granularity decides whether any of this is useful. Our first replay
certified whole decode steps against the step's worst sampled entry; one
large-witness entry then forced the entire step to fall back, and coverage was
$0\%$ at every useful threshold. Local fallback must be local: certifying
\emph{entries} makes the per-step TV bound hold by construction, and replaying
$12.4$M entry reads across the $21$ sparse-selection layers gives the
trade-off curve---$80.1\%$ compressed-entry retention at a vacuous per-step
bound of $1.54$, or $\mathrm{TV}\le 0.86$ at $29.3\%$ retention
(we say retention rather than certified coverage: quoting the retained fraction
alone would hide that its per-step bound is vacuous). On the int8
re-quantisation witness the distribution is concentrated with no heavy tail to
cut, so a non-vacuous bound and $80\%$ retention cannot hold simultaneously:
the witness, not the calculus, is the binding constraint. In this offline replay
all certificates are deterministic and zero budget is spent
($\Sigma\delta=0$); the write-time experiment below spends the probabilistic
budget for real. The
final two pieces touch the numeric path, and a controlled experiment lands both at
write granularity---with one correction a stability control forced on our own
earlier reading. Compression is made real by FP8 mantissa masking (a genuine
coarsening of every stored window entry's value grid, though the byte count is
unchanged: this is precision coarsening, not physical compression), and the
per-entry certificate decides, entry by entry, whether the coarsened bytes are
written or the exact bytes are kept: fallback executes in the numeric path with
zero page-in by construction. The policy accepts $48.0\%$ of entries for
masking (a policy acceptance rate, not a retained-bytes figure).

Probabilistic budget is spent with \emph{pre-authorization} semantics:
a prior radius authorizes an entry before any random bits are drawn, a
write-side audit counts realized violations ($0$ of $147{,}208$ authorized
entries), and the price of authorizing before the draw is quantified rather
than hidden ($6.2\times$ fewer entries than post-hoc gating accepts;
mechanics in Appendix~\ref{app:ledgermech}).

The read-side per-request ledger and the write-side policy run in the \emph{same}
serving process over the same requests, and the loop now closes on the
\emph{same physical entries}: the certified write acts on the compressed-page
pool---the buffer the read-side ledger dequantizes---and a per-slot status map
(unwritten / masked / kept-exact) links every policy decision to its later
reads. Over eight serial user requests, $3{,}936{,}205$ of $7{,}750{,}144$
read-side entry checks landed on slots the policy had masked, with zero reads
of unwritten slots. Slot reuse is handled by \emph{owner-tagged slot state}
(with a per-rewrite generation counter): the
compressed pool has no separate free path, so a reused slot is necessarily
rewritten through the certified writer before it can be read again---the
status map stays fresh by construction---and each write records the writing
request's UID. Under real reuse pressure ($11{,}763$ owner-change events from
serial page recycling) the per-request closure stays exact: distinct prompts
read zero foreign-owned entries, while a deliberately repeated prompt reads
$94\%$ of its entries from the prefix cache ($974{,}386$ of the run's
$1{,}063{,}892$ foreign-owned reads; the remainder trace to server warmup
batches that carry no request identity) and every one is attributed to
its writing request rather than miscounted into the reader's closure
(write-side $\delta$ is charged to the owner; read-side checks are
deterministic, so sharing does not disturb budget semantics). 
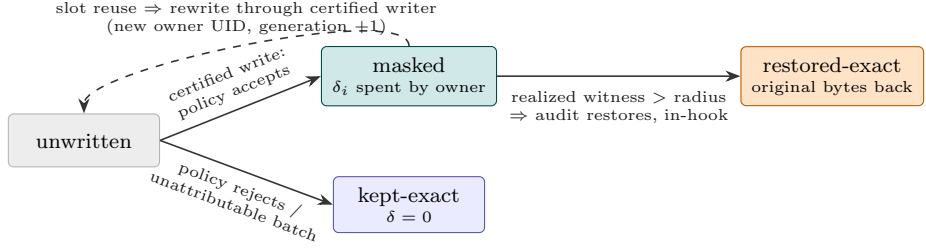
\begin{figure}[t]
\centering
\begin{tikzpicture}[font=\scriptsize, node distance=9mm,
  st/.style={draw, rounded corners=2pt, align=center, inner sep=4pt,
             minimum height=7mm, minimum width=20mm},
  un/.style={st, fill=gray!14, draw=gray!60},
  mk/.style={st, fill=teal!18, draw=teal!55!black},
  ke/.style={st, fill=blue!8, draw=blue!35!gray},
  re/.style={st, fill=orange!22, draw=orange!70!black},
  arr/.style={-{Stealth[length=2mm]}, semithick, gray!40!black},
  lab/.style={font=\tiny, align=center, gray!20!black}]

\node[un] (u) at (0, 0)      {unwritten};
\node[mk] (m) at (4.3, 0.85) {masked\\[-2pt]\tiny $\delta_i$ spent by owner};
\node[ke] (k) at (4.3, -0.85){kept-exact\\[-2pt]\tiny $\delta = 0$};
\node[re] (r) at (9.9, 0.85) {restored-exact\\[-2pt]\tiny original bytes back};

\draw[arr] (u.east) -- (m.west)
  node[lab, midway, above, sloped]{certified write:\\policy accepts};
\draw[arr] (u.east) -- (k.west)
  node[lab, midway, below, sloped]{policy rejects /\\unattributable batch};
\draw[arr] (m.east) -- (r.west)
  node[lab, midway, below=0.5mm]{realized witness $>$ radius\\$\Rightarrow$ audit restores, in-hook};
\draw[arr, dashed] (m.north) .. controls +(0,0.6) and +(0,0.6) .. (u.north)
  node[lab, midway, above]{slot reuse $\Rightarrow$ rewrite through certified
  writer\\(new owner UID, generation $+1$)};
\end{tikzpicture}
\caption{\textbf{The ledger's slot-state machine.} Admission into
\emph{masked} is probabilistic (it spends the owner's $\delta_i$);
enforcement is deterministic---a violating entry is restored to exact
bytes in the same hook call, before any shared read. Reused slots
re-enter only through the certified writer, with fresh owner and
generation.}
\label{fig:slotstate}
\end{figure}

Four runtime
invariants make shared-read semantics fail-closed rather than
observational (Figure~\ref{fig:slotstate})---write-audit before any shared read, explicit slot state,
on-the-spot byte restoration (admission is \emph{probabilistic},
enforcement is \emph{deterministic}), and once-per-write budget spend
(\texttt{shared\_read\_sound}); the precise statements are
Appendix~\ref{app:ledgermech}. Mid-request
eviction workloads remain uncovered. Request identity is a never-reused monotone UID detected on the
write path itself, so account boundaries coincide with true request
boundaries; each request carries an independent telescoping budget (the
eight user accounts each stay under budget, maximum spend
$\Sigma\delta = 0.009982 \le \delta_{\mathrm{req}} = 0.01$), and the
stochastic-rounding stream is per-(nonce, UID, layer, counter) and
rank-invariant across the eight replicas
(mechanics in Appendix~\ref{app:ledgermech}).
Pre-draw authorization audited $0$ violations in $120{,}661$ authorized
entries on this pool. Concurrency is covered by per-row identity: within a
mixed batch every entry is attributed to its request via the compressor's
plan alignment (decode plans are batch-row-aligned; prefill plans are
per-query-token, segmented by the prefix sum of extend lengths), each account
keeps its own budget counter and random stream, and under eight
simultaneously-submitted requests every account independently satisfies its
budget (maximum spend $\Sigma\delta = 0.009984 \le \delta_{\mathrm{req}}$)
with the owner ledger reporting
\emph{zero} foreign-owned reads across $8{,}228{,}864$ checks. That owner
ledger doubles as a runtime guard on attribution itself: its first concurrent
run flagged $3.4\%$ foreign reads and traced them to a namespace collision
between the two compressed-pool families---invisible under serial traffic,
where a single active request masks ownership churn---which is precisely the
class of accounting error a purely serial evaluation cannot surface. On output quality we report the
control first: greedy decoding across two independent server restarts with
\emph{no} intervention already disagrees on $2$ of $8$ diverse prompts.
Against that stability floor, the policy and dithered arms' output differences
from baseline are at or near the floor at $n{=}8$ and cannot be resolved; the
mask-everything control still diverges grossly, so the intervention is real, and
we describe the gated arms as \emph{consistent with} a protective effect rather
than demonstrating one. We use the word executable for exactly this scope---
write-time policy with pre-authorized probabilistic budget, plus read-side
per-request accounting in the same run---and quality claims stay at the
eight-prompt directional level.

\subsection{Upgrading the certified object}
The witness being the binding constraint, we changed the certified object itself:
from a worst-case norm to the exact conditional distribution of stochastic
rounding. An offline adjudication over raw captured traces (queries, entry bytes
and scales for all $43$ layers; $64$ rounding realizations per entry as
ground truth) compared four certificate families against pre-registered gates.
The centred bound---softmax is shift-invariant, so only the oscillation of the
score error matters, and the two-point extremal argument gives the sharp constant
\begin{equation}
\mathrm{TV} \;\le\; \tanh\!\bigl(\mathrm{osc}/4\bigr)
\label{eq:tanhbound}
\end{equation}
---moves non-vacuous layers from
$0/43$ to $43/43$ on this object. Tightness is measured against the
matching realized object per rounding realization (a review corrected our first
metric, which divided a centred bound by an uncentred realization): the centred
oscillation bound sits $5.2\times$ above realized oscillation, and the exact-MGF
per-entry radius $2.1\times$ above realized $|\Delta s|$ under a joint
$\delta/E$ budget, against $15.8\times$ for the current-form witness bound. Three of our own bugs were
caught by the adjudicator's soundness self-check (radii must never be beaten by
realizations): a two-sided Chernoff combined with $\min$ instead of
$\max$; a deterministic shift from top-of-range clamping that is not random at
all; and a zero-weight branch producing $\log 0$ that a clamp silently turned
into fake perfection. One gate failed and is honoured: no family reaches
$\tau{=}0.2$ retention at two-bit coarsening, so the heavy fused-kernel stage
is skipped by the pre-registered rule.

Deployed at write time---where the pre-quantization value is in hand, so the
rounding position $\theta$ costs no metadata---true stochastic rounding
(conditionally exact-zero mean) with a \emph{variance-aware two-point
bounded-difference radius}---exact two-point variance
$\theta(1{-}\theta)\Delta^2$ and exact clamp shift, composed through Jensen
and a Hoeffding tail. All three components of this radius are
machine-checked on the finite product space with the constants the
implementation uses---no measure theory anywhere, because the randomness
is finitely many discrete draws:
\begin{align}
\text{tail:}\quad&
  F'' = \frac{\theta(1-\theta)E_1E_2}{g^2} \le \tfrac14
  &&\text{\scriptsize two-point Hoeffding, sub-Gaussian }
    \scriptstyle\Delta^2/4\label{eq:radtail}\\
\text{witness:}\quad&
  \text{bounded-difference MGF on the band witness}
  &&\text{\scriptsize finite-}\scriptstyle\Omega\,
    \text{\scriptsize McDiarmid, by induction}\label{eq:radwit}\\
\text{mean:}\quad&
  \mathbb{E}\sum_b\|d_b\| \le
  \sum_b\sqrt{\textstyle\sum_c \theta(1-\theta)\Delta_c^2}
  &&\text{\scriptsize recursive Jensen step}\label{eq:radmean}
\end{align} \begin{figure}[t]\centering
\includegraphics[width=0.98\textwidth]{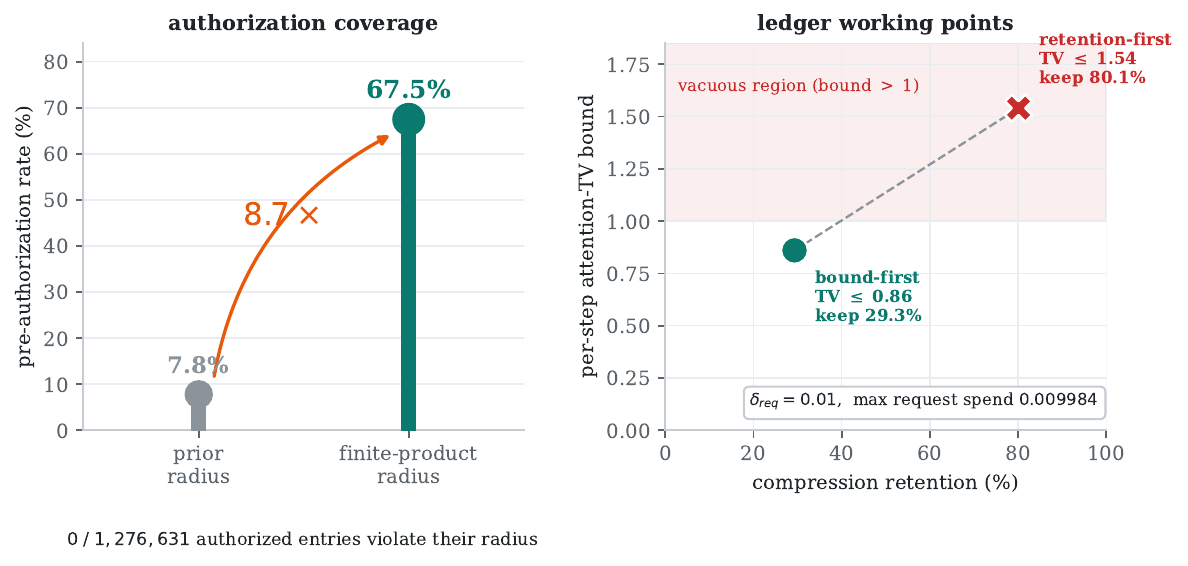}
\caption{\textbf{The certified object, quantitatively.} Left: the
finite-product radius lifts pre-authorization coverage from $7.8\%$ to
$67.5\%$ at the same threshold ($0$ of $1{,}276{,}631$ authorized
entries violate). Right: the ledger's two working points---retention-first
keeps $80.1\%$ but its per-step bound ($1.54$) is vacuous; bound-first
certifies TV $\le 0.86$ at $29.3\%$ retention. Max request spend
$0.009984 \le \delta_{req}=0.01$.}
\label{fig:math}
\end{figure}
The radius raises the pre-authorization rate at the same threshold from
$7.8\%$ to $67.5\%$ ($8.7\times$), with $0$ of
$1{,}276{,}631$ authorized entries violating their radius. For the budget,
a finite-$\Omega$ Ville inequality---provable by a three-line induction on
hitting probabilities, with no measure theory, because the randomness is finitely
many discrete draws---gives an anytime-valid e-process, and we now run one
\emph{on the same object as the ledger}: each authorized entry contributes
\begin{equation}
g_e \;=\; \exp\!\bigl(\lambda_e(W_e - m_e) - \lambda_e^2 C_e/8\bigr),
\qquad \lambda_e = 4\kappa/\sqrt{C_e},\; \kappa = 0.5
\text{ pre-registered},
\end{equation}
built from its realized witness $W_e$ against its pre-draw mean term
$m_e$; the supermartingale condition
$\mathbb{E}[g_e \mid \mathcal{F}] \le 1$ is itself machine-checked (it
composes the McDiarmid MGF bound with the mean-term lemma).
The sentinel runs at two scopes: per-request accounts, and a \emph{global}
process that never resets across requests---the continuously-running,
whole-service form. Over $120{,}661$ authorized-entry factors spanning all
requests, the global process never approaches its crossing threshold:
\begin{equation}
\sup_t\,\log M_t \;=\; -0.82
\;\;\ll\;\;
\ln(1/\delta_e) \;=\; 4.61
\label{eq:eprocesscheck}
\end{equation}
(final value $-76{,}819$; we report the post-update maximum rather than
the initialization value, which would overstate the evidence): an anytime-valid
check that the stochastic-rounding radius assumptions are not contradicted by
live traffic. Its semantics are model validation---a drift sentinel on the
same entries at the same instants---complementing, not replacing, the
per-entry union budget that authorization consumes. Separately, on the two-sided
cumulative object with $33{,}407$ events the e-process radius is $390\times$
tighter than summing telescoped per-event radii; the two certify
\emph{different failure events} (``any per-entry radius is beaten'' versus the
signed cumulative sum), and we report that magnitude comparison under that
explicit caveat. The unfused write-path
implementation costs $1.20\times$ in per-request wall clock (graphs-off
regime, serial requests); the fused-kernel stage that would reduce
it was skipped by the pre-registered gate, and physical bytes are unchanged
throughout---precision coarsening is not physical compression.

\subsection{Corruption sentinel}
Over $144$ baseline verification passes we observed zero false alarms; this is not a
false-positive rate of zero, and by
\begin{equation}
(1-p)^n \;\le\; e^{-np}
\label{eq:sentinelbound}
\end{equation}
the one-sided $95\%$
upper bound on the rate is $2.06\%$. In the injection experiment, $2$ of $4$
corrupted slots were located. Lemma~\ref{lem:sentinel} explains why this is expected
rather than a limitation---with one correction this revision makes explicit: the
implementation samples \emph{with} replacement, so the model must too, and our
earlier hypergeometric analysis was optimistic about the implementation it described.
With replacement, an effective post-injection window of $n\approx 17$ rounds
sampling $128$ of $1024$ slots gives a per-block detection probability of
$0.881$, and the probability of finding at most two of four is $0.072$.
Detection latency is a budget, not a capability: reaching $99\%$ confidence for a
single corrupted slot takes $148/74/37/19$ rounds at $r=32/64/128/256$.

\subsection{Overhead}
\begin{figure}[t]\centering
\includegraphics[width=0.95\textwidth]{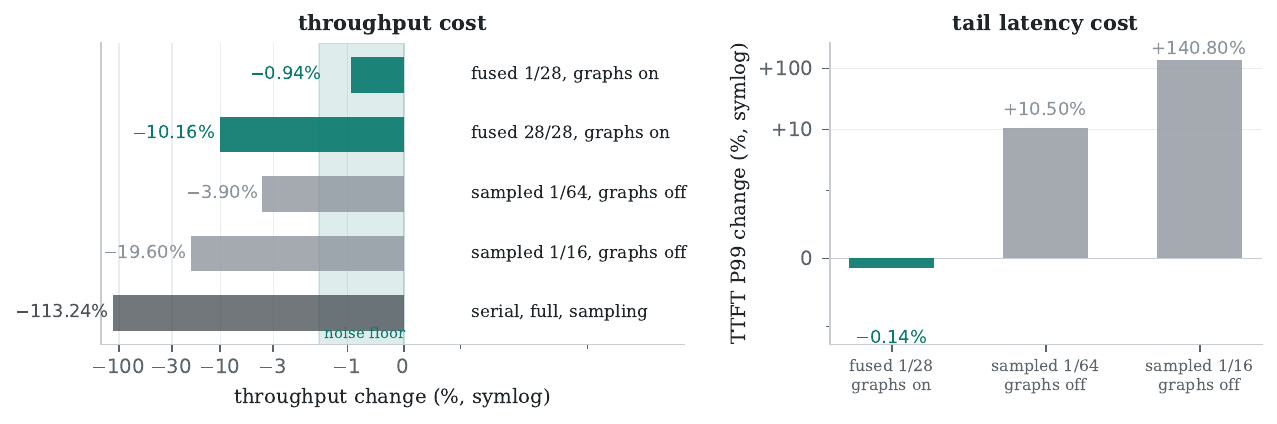}
\caption{\textbf{Resident probe cost.} Under CUDA graphs the
fused meter at one of 28 layers costs $-0.94\%$ throughput ($-0.14\%$
TTFT P99), inside the noise floor; full coverage ($-10.16\%$) remains a
diagnostic mode. Graphs-off sampling points shown for cross-regime comparison.}
\label{fig:overhead}
\end{figure}
\label{sec:overhead}
Three measurement regimes are necessary, and the always-on claim rests on the last. Serially, with one request in flight and CUDA graphs
disabled, the probes cost $+113.24\%$ at full sampling---the worst-case
amplification, with no other work to hide host-side cost behind. Under concurrency
the picture separates by sampling rate: at one sample per $64$ calls the cost is
$-3.9\%$ throughput and $+10.5\%$ TTFT P99, both inside the per-setting measurement
noise, whereas one sample per $16$ calls costs $-19.6\%$ and $+140.8\%$, clearly
outside it. Two methodological points follow. The noise floor must be computed per
setting; taking a global maximum would classify the $16$-sample degradation as noise.
And tail latency, not throughput, is the criterion for always-on operation: the two
settings are indistinguishable on throughput alone.

The third regime is graphs \emph{enabled}, and it is the configuration that
justifies the word always-on: instrumenting one declared layer with the fused
Triton meter costs $-0.94\%$ throughput and $-0.14\%$ TTFT P99 against an
identical probe-off baseline, both inside the per-concurrency noise floor, with
replay-time accumulation and declared-versus-observed coverage verified by the
acceptance gate. How this configuration was reached---including the failure of
call-rate sampling under graph capture and the acceptance-gate hole it exposed---is
Appendix~\ref{app:overhead}; full-coverage instrumentation under graphs remains
a diagnostic mode at $-10.16\%$.

\begin{table}[t]\centering\small
\begin{tabular}{@{}>{\raggedright\arraybackslash}p{0.24\textwidth}>{\raggedright\arraybackslash}p{0.26\textwidth}>{\raggedright\arraybackslash}p{0.17\textwidth}>{\raggedright\arraybackslash}p{0.24\textwidth}@{}}
\toprule
Claim & Evidence & Strength & Not yet shown\\
\midrule
Request-level risk budget &
Lean-checked ledger + 8-request serving replay &
certified &
cross-layer error propagation\\
Eviction-free packed consumption &
three independent 12-doc rounds, exact &
empirical exactness &
eviction / slot-reuse safety\\
Ring write/finalize/recycle chain &
machine-adjudicated concurrent gate, zero violations &
diagnostic chain &
non-resident recovery\\
Low-overhead resident probe &
graph-on serving benchmark &
one layer of 28 &
full-coverage always-on\\
\bottomrule
\end{tabular}
\caption{\textbf{Claim boundaries.} What is asserted, on what evidence, at
which tier, and what is explicitly not asserted; each row's numbers appear
in the text with their measurement regimes.}
\label{tab:boundary}
\end{table}

\section{A served DeepSeek-V4 packed prototype: diagnosis to consumption}
\label{sec:v4}
The machinery of \S\ref{sec:eval} was built to observe; this section is what
happened when it was pointed at a served DeepSeek-V4 stack carrying a packed
compressed-KV prototype: a certificate-guided format exploration, a silent
corruption localized to a structural boundary, the packed path served end to
end, and a six-path serving-tax taxonomy---every verdict generated from run
artifacts, not summarized by hand.

\subsection{Case study: certificate-guided low-bit format exploration}
A boundary-discovery exercise drove an aggressive low-bit format study on
V4-Flash's compressed pools (block-Hadamard INT6/INT4, a fused unpack kernel
$3\times$ faster than the production FP8 dequantizer, a real packed pool in
serving); the instrument falsified two of our own design assumptions and
scoped precisely what a served consumption claim would require. The full
account is Appendix~\ref{app:casestudy}; its open item---served packed
consumption---is what \S\ref{sec:served} then closes and stresses.

\subsection{Closing the boundary: generation staleness, not quantization}
\label{sec:generation}
The scoped follow-up closed the boundary, and the closure is itself a
result about what the type system must carry: a $95$-point retrieval
collapse that radix caching had been masking was discriminated down to
\emph{generation staleness}---slot reuse with no invalidation call
anywhere in the codebase---rather than quantization (the quantizer's
intrinsic relative error is $0.0245$, unreachable from the observed
readings), and unconditional write-site invalidation restored accuracy
with a two-sided causal confirmation ($1.000$ vs.\ $0.200$ under
$4$-way concurrency, six machine-checked criteria). The full
discrimination narrative is Appendix~\ref{app:staleness}.

Two lessons feed back into the contract layer. First, the \emph{observation
site} is part of a metric's type. The write-site shadow check---mathematically
identical to the decode-side one---had zero discriminative power for this
failure ($0.967$ vs.\ $1.008$ across fix/control arms) because the pool has
two writers and the site compares a decode-time snapshot against a write-time
source: a cross-generation comparison by construction. The same metric at the
consumption point (what the kernel is about to read, sampled $1/32$, before
any refill) separates the arms by $36\times$: $0.0246$---matching the
$0.0245$ intrinsic within half a percent, reproduced across six independent
runs---against $0.882$. A bridge between metrics must therefore check not
only units but \emph{where} the quantity is measured. Second,
$(\mathrm{slot}, \mathrm{generation})$ is a contract \emph{premise}, not a
logging field: the refinement obligation is that a consumed entry's
generation equals the generation at authorization, and write-site
invalidation is its enforcement half.

The claims withheld above were then partially discharged---served
consumption at parity, structural pool replacement quality-neutral,
mechanism-level memory reduction with measured/accounting agreement of
$1.7\%$---with one explicitly not discharged: the ring-staged configuration
degraded retrieval while every fail-loud surface stayed green (the full
discharge ledger, including the out-of-bounds addressing finding, is
Appendix~\ref{app:discharge}).
Both open items were subsequently discharged, and
\S\ref{sec:served} records the closure and its measurement regime.

\subsection{Serving the packed format: from diagnosis to consumption}
\label{sec:served}
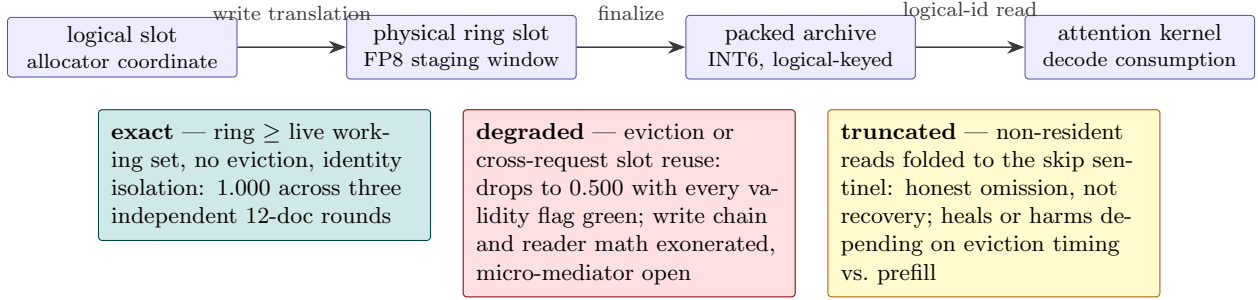
\begin{figure}[t]
\centering
\resizebox{\linewidth}{!}{%
\begin{tikzpicture}[font=\footnotesize,
  st/.style={draw, rounded corners=1.5pt, align=center, inner sep=3.5pt,
             minimum height=8mm},
  lc/.style={st, fill=blue!7, draw=blue!35!gray, text width=29mm},
  card/.style={st, align=left, text width=42mm, inner sep=5pt},
  ok/.style={card, fill=teal!18, draw=teal!55!black},
  bad/.style={card, fill=red!12, draw=red!55!black},
  mid/.style={card, fill=yellow!24, draw=orange!70!black},
  arr/.style={-{Stealth[length=2.4mm]}, semithick, gray!45!black}]

\node[lc] (log)  at (0, 0)     {logical slot\\[-1pt]{\scriptsize allocator coordinate}};
\node[lc] (ring) at (4.65, 0)  {physical ring slot\\[-1pt]{\scriptsize FP8 staging window}};
\node[lc] (arch) at (9.3, 0)   {packed archive\\[-1pt]{\scriptsize INT6, logical-keyed}};
\node[lc] (kern) at (13.95, 0) {attention kernel\\[-1pt]{\scriptsize decode consumption}};
\draw[arr] (log.east) -- (ring.west)
  node[midway, above=2.4mm]{\scriptsize write translation};
\draw[arr] (ring.east) -- (arch.west)
  node[midway, above=2.4mm]{\scriptsize finalize};
\draw[arr] (arch.east) -- (kern.west)
  node[midway, above=2.4mm]{\scriptsize logical-id read};

\node[ok, anchor=north]  (g) at (1.95, -0.85)
  {\textbf{exact} --- ring $\geq$ live working set, no
   eviction, identity isolation: $1.000$ across three
   independent 12-doc rounds};
\node[bad, anchor=north] (r) at (6.95, -0.85)
  {\textbf{degraded} --- eviction or cross-request slot
   reuse: drops to $0.500$ with every validity flag green;
   write chain and reader math exonerated,
   micro-mediator open};
\node[mid, anchor=north] (y) at (11.95, -0.85)
  {\textbf{truncated} --- non-resident reads folded to the
   skip sentinel: honest omission, not recovery; heals or
   harms depending on eviction timing vs.\ prefill};
\end{tikzpicture}}
\caption{\textbf{Compressed-entry lifecycle and runtime regimes.} The exact
regime is certified; degradation correlates with eviction and slot reuse
while the write chain and audited readers are exonerated (mediator open,
\S\ref{sec:served}); the sentinel fold is a timing-dependent mitigation.
The contract lesson: a replaced KV representation redefines coordinate,
generation, and materialization semantics---all three belong in the
runtime contract.}
\label{fig:v4mech}
\end{figure}

\begin{figure}[t]\centering
\includegraphics[width=0.86\linewidth]{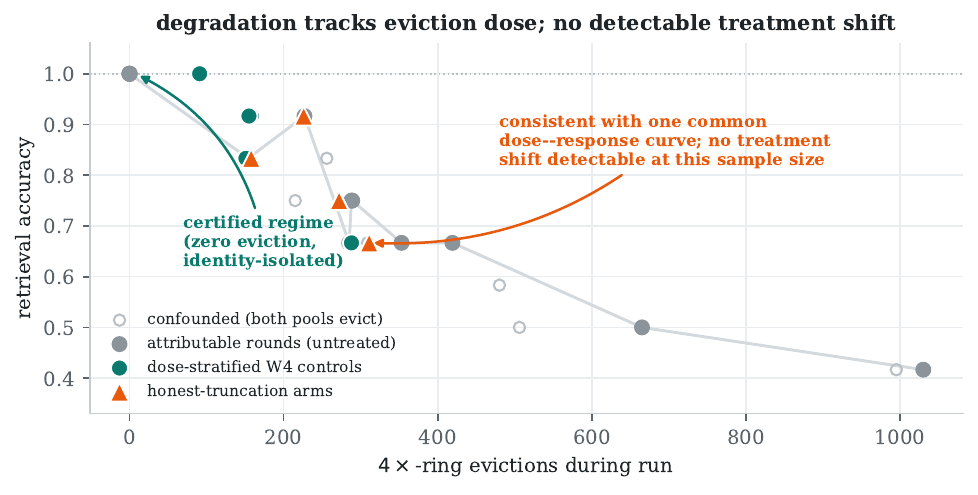}
\caption{\textbf{Eviction dose governs quality.} Historical
attributable rounds (gray; open = dose-confounded, context only),
dose-stratified twelve-document controls (teal), and honest-truncation
arms (orange) are consistent with one common dose--response relationship; no
treatment shift is detectable at the present sample size. The zero-eviction cluster is the certified regime.
Every point is read from its run artifact.}
\label{fig:dose}
\end{figure}

The preceding subsection ended with two open claims---the ring lifecycle
unproven, the $4\times$ pool untouched. The diagnostic ring chain and the
physical pool-integration gates are now closed by machine-gated runs
(per-\emph{(layer, slot, epoch)} coverage with zero out-of-generation
reads across eight ranks; an end-to-end packed INT6 replacement of the
$4\times$ pool with measured $2{,}108$\,MiB/GPU HBM reduction against a
tensor-derived byte model; in-graph packed consumption at exact retrieval
parity), while eviction-safe production consumption remains open. What
turned a quality regression into a structural boundary was a
machine-adjudicated discrimination campaign---every comparison a
controlled pair, every verdict a run artifact, two of our own confounded
inferences rejected by the adjudication calculus on the way. Its shape:
a per-pool dissociation pinned the harm to the $4\times$ ring's
eviction (dose-linked, reproduced across rounds); write-side forensics
($16{,}942{,}020$ bake-versus-store row checks, zero mismatches)
exonerated the write chain; content oracles judged against calibrated
repeat-pair noise floors proved the persistent stores and the entire
prefill computation arm-invariant; and an intervention that folds
non-resident reads to the kernel's skip sentinel restores the eviction
microcosm from $0.500$ to $1.000$ under matched eviction and call
counts---honest truncation, effective but timing-sensitive, a
mitigation rather than a repair. At twelve-document concurrency,
eviction dose itself is scheduling-stochastic, and the truncated
and untreated arms are consistent with one common dose--response
relationship---no treatment shift is detectable at the present sample
size (a quick regression reads $+3.35$pp with an interval spanning
zero)--- so the certified
operating point is structural, not parametric
(Figure~\ref{fig:v4mech}, and quantitatively
Figure~\ref{fig:dose}: thirty-two runs from three series are consistent
with one common dose--response relationship). The exclusion list is now long and
two-sided: beyond the write chain and the persistent stores, a
materialized reference reader---the instrument-verified dequantization
path with reference attention mathematics, executed $1{,}430$ times in a
matched arm---reproduces the same $0.500$, so the production kernel's
inline unpack and the reader mathematics are exonerated as well. What
remains is a stable structural association---every observed failure occurs in an
eviction or slot-reuse regime, while eviction alone does not force
failure (one $91$-eviction control run scores $1.000$)--- whose microscopic mediator is open, with the
remaining candidate surface in selection metadata, index semantics, or
their timing state; we record it as an open question rather than
narrate beyond the evidence. The full round-by-round adjudication log,
with every number read mechanically from run artifacts, is
Appendix~\ref{app:campaign}. What is proven and
production-relevant is unchanged: the zero-eviction,
identity-isolated configuration is exact across three independent
twelve-document rounds, and it is the only configuration this campaign
certifies.

\subsection{Path-contract evaluation}
\label{sec:pathcontract}
Bringing a replaced KV representation into a production stack is not one
compatibility question but six, and the value of a runtime contract is that
it answers each with a different failure type. We evaluated the packed
stage against six serving paths and report a five-level taxonomy
(Table~\ref{tab:sixpath})
---\textsc{pass} / \textsc{fail-method} / \textsc{blocked-upstream} /
\textsc{partial} / \textsc{not-measured}, of which four occur across these
six paths---generated mechanically from the
run artifacts rather than summarized by hand.
\newcommand{\ok}{\textcolor{teal!60!black}{$\checkmark$}}
\newcommand{\bad}{\textcolor{red!60!black}{$\times$}}
\newcommand{\pt}{\textcolor{orange!80!black}{$\triangle$}}
\begin{table}[t]\centering\small
\setlength{\tabcolsep}{3.5pt}
\begin{tabular}{@{}lcccccc l@{}}
\toprule
Path & Capt. & Id/gen & Packed & Evict & Xfer & Qual. & Verdict\\
\midrule
CUDA graphs & \ok & \ok & \ok & \pt & --- & \pt & \textsc{partial}\\
TP repeat & \ok & \ok & \ok & --- & --- & \ok & \textsc{pass}\\
Radix cache & \ok & \bad & --- & \bad & --- & \bad & \textsc{fail-m}\\
MTP (spec.\ decode) & \ok & \ok & \ok & \ok$^*$ & --- & \ok & \textsc{pass}\\
PD disagg. & \ok & \ok & \bad & --- & \bad & \bad & \textsc{partial}\\
HiSparse & --- & --- & --- & --- & --- & --- & \textsc{blocked}\\
\bottomrule
\end{tabular}
\caption{\textbf{Path-contract matrix}, generated from run
artifacts. Cells mark which contract layer each path violates
($\triangle$ = holds at a stated caliber; --- = n/a or
unmeasurable; $^*$ = zero-eviction caliber, machine-annotated);
per-verdict evidence in \S\ref{sec:pathcontract}.}
\label{tab:sixpath}
\end{table}
 The matrix carries the per-dimension verdicts;
prose here keeps only what each row \emph{teaches}
(per-path evidence and regimes in Appendix~\ref{app:paths}).
CUDA-graph serving is exact in the eviction-free regime and
regime-limited beyond it ($0.75$ at twelve-document concurrency);
radix prefix caching fails with the method reproducibly ($0.667$ vs.\
$1.000$ native, two independent rounds), so the installer refuses the
combination loudly; speculative decoding passes on a two-seed
zero-eviction rerun ($1.000$ on both arms, both seeds) after the
adjudication chain attributed its historical residual to the eviction
mechanism. Prefill/decode disaggregation yields the sharpest lesson: the
native arm passes exactly ($1.000$; the needle lives in the prefill
worker's context, so a correct decode-side answer certifies a real KV
transfer), while the packed arm returns $0.000$ with \emph{zero} transport
or server errors---transfer completed, kernels ran, service stayed
up---because the transfer engine moves bytes by the pool-layout contract
that the packed stage redefines. A run can be structurally green and
semantically wrong; that is precisely the gap a runtime quality oracle
exists to close. The same oracle then sharpened the eviction-free claim
itself: in a matched benchmark round in the working-set-ring regime, a
retrieval battery run \emph{after} a 32-request load scored $0.917$ with
zero evictions---the missed document answered with \emph{another
request's} needle code, because freed logical slots were reused while
their ring pages stayed resident, so the ring served the previous
owner's rows without any eviction. Zero eviction is therefore necessary
but not sufficient; eviction and cross-request slot reuse are two
triggers of one structural exposure; the sentinel-fold intervention
removes the exposed reads at microcosm scale (a timing-sensitive
mitigation, \S\ref{sec:served}), and eviction-safe consumption with a
verified mediator account remains future work. The remaining path (hierarchical sparsity) is blocked
upstream: the unmodified baseline crashes identically with and without
graphs, so the path is unmeasurable in this build; the crash trace is
preserved as an incident artifact so the classification is auditable
rather than asserted. On cost we claim
nothing yet: the staged consumer serves at $0.884\times$ native throughput
and $1.131\times$ cost per output token in its unoptimized
materialize-per-call form, and the fused-kernel arm's $0.238\times$ /
$4.201\times$ figures include per-call host-side auditing scaffolding that
production operation would remove; these are upper bounds on overhead, not
a benefit statement. On quality in the same regime, a
teacher-forced paired perplexity round (24 natural+code documents,
$98{,}568$ tokens, tp4-matched arms, with bidirectional path-activation
evidence: the packed arm records $1{,}112$/$1{,}040$ packed
consumptions on the two pools; the base arm carries no packed switches,
adapters, or consumption counters, with absence recorded in its
manifest) reads $1.3496$ vs.\ $1.3495$ ($-0.00\%$; successive matched
pairs read $+0.05\%$ and $-0.002\%$, within run-to-run jitter): the
packed consumption path is perplexity-neutral, so the cost story is
not hiding a quality tax. The matched final-table round quantifies the
unoptimized path end to end in the working-set-ring regime: $26.8$
versus $173.4$ output tokens/s against the native arm on a 32-request
benchmark, with the enlarged rings erasing the memory saving in this
regime---an honest cost statement, with the benefit claim still
withheld until eviction-safe consumption is demonstrated.

\section{Limitations}
\label{sec:limits}
The sampled-probe overhead measurements of Section~\ref{sec:eval} are taken
with CUDA graphs disabled, so their absolute latencies do not represent
graph-enabled production and only the relative deltas are meaningful; the
graph-enabled regime is measured separately (\S\ref{sec:overhead}) and
carries its own scope. The probe itself, however, does run under
graph capture, and we state precisely what we verified, because ``the server starts''
is far too weak a test: graph replay executes no Python at all, so a probe whose
accumulation quietly stopped would look identical to one that worked. Liveness under
capture is verified by exact count arithmetic (host and device counters
grow by precisely the predicted call and replay totals), and the eight
silent-corruption trap classes this took---four on the probe side, four
rediscovered on the packed write side, each invalidating capture rather
than crashing---are catalogued with their regression tests in
Appendix~\ref{app:overhead}. The path-contract matrix of
\S\ref{sec:pathcontract} is likewise a bounded claim: one model, one
serving stack version, small retrieval batteries, largely single-seed; it
demonstrates the \emph{taxonomy}---which paths fail, in which of five
distinct ways---not production coverage, and no end-to-end cost benefit is
claimed anywhere in this paper.

The always-on result of \S\ref{sec:overhead} (one declared layer,
fused meter, $-0.94\%$/$-0.14\%$, Figure~\ref{fig:overhead}) took two
changes---replay-constant sampling decisions and a fused Triton meter
$24.2\times$ cheaper per invocation ($9.3e-07$ max relative error against
the reference)---whose engineering account, including why naive sampling
silently vanishes from captured graphs and how a lifetime-coverage gate
passed on a dead probe, is Appendix~\ref{app:overhead}. We state the scope precisely: what is demonstrated is always-on
observation of \emph{a declared subset}, not of every layer. Full $28$-layer
coverage under graphs still costs $-10.16\%$ throughput even fused, which is
far outside the noise floor, so contract readings that need every layer remain a
diagnostic mode rather than a resident one. The
storage term is now denominated in total variation with both window and
compressed-page entries witnessed on the sparse-selection layers, and the selection
term crosses metrics through a measured ratio; two gaps remain open, and we name
them. The page side is now witnessed on every layer that has one: of the
$43$ layers, $21$ select over $4\times$-compressed pages, $20$
attend densely over $128\times$-compressed pages (no indexer, hence no
selection term), and the first two are uncompressed by architecture; the
coarse-compression pages turn out to carry a witness about half the size of the
sparse-selection pages' ($0.523$ vs.\ $1.045$ max), and their median
worst-case per-layer TV bound is $0.976$ against the sparse-selection
layers' $1.339$: the harder-compressed layers are, counterintuitively, the
easier ones to certify. And the per-layer TV bounds do not compose
across layers---each layer's attention is a different distribution, and we claim no
inter-layer bridge---so the request-level number is an aggregate budget, and a
vacuous one. The empirical bridge ratio is measured on truncation sets and applied
to perturbation swap sets, a homogeneity assumption the tier system prices in by
holding the composed chain at \emph{empirical}. The
recurrent-state aggregate does not retain channel identity, as noted. The sentinel
guarantee assumes corrupted slots are independent of the sampling. Finally, six evaluated model
configurations from one serving stack are not a substitute for independent
replication. The release boundary is stated plainly: the public repository
reproduces every number, figure and theorem from shipped artifacts and
sources, while the architecture-specific injection implementation (hook
anchors, adapters, kernels) remains private---serving-level reruns
therefore require the platform, and the paper's verifiable surface is the
artifact-and-guard chain, not a rerun claim.

\section{Related Work}
\label{sec:related}
\textbf{Runtime certificates for compressed attention.} Prior work bounds the
distortion introduced by KV-cache quantisation---per-channel and outlier-aware
schemes such as KIVI~\cite{kivi} and KVQuant~\cite{kvquant}---and uses the bound
to gate or repair at decode time. That line fixes the memory object---a dense post-RoPE KV cache---and
derives one bound for it. Our concern is orthogonal: we do not propose a tighter
bound for that object, we ask what remains true when the object itself changes, and
what has to be supplied for a new object to be admitted.

\textbf{Compression for latent and sparse architectures.} Latent caches
(MLA~\cite{deepseekv2}), learned sparse selectors~\cite{nsa}, pooled
hierarchies and linear-attention states~\cite{gla}, and token-selection or
eviction schemes (H2O~\cite{h2o}, SnapKV~\cite{snapkv}, attention
sinks~\cite{streamingllm}, GEAR~\cite{gear}) each report end-task scores
computed offline against an uncompressed reference---exactly the quantity a
serving system does not have; an early version of our own composition used
that true residual as a bound before the tier registry rejected it as
unavailable at runtime.

\textbf{Systems observability.} Production serving stacks such as
vLLM~\cite{vllm} and SGLang~\cite{sglang} expose throughput, latency and memory,
and hierarchical caches expose hit rates; none of these observes whether
the retained state is still faithful. The incident that motivates
\S\ref{sec:intro} was invisible to all of them. The sentinel of
Lemma~\ref{lem:sentinel} is deliberately in the shape systems people already
use---a sampled check with a stated detection latency and a tunable budget---rather
than a new guarantee requiring a new abstraction.

\textbf{Anytime-valid inference and concentration.} The per-entry radius is
classical concentration~\cite{hoeffding,mcdiarmid} deployed
\emph{prospectively} (declared before the draw, funded by a budget that
telescopes soundly at unknown length), and the drift sentinel is an
e-process~\cite{ramdas-savi,ville,howard-chernoff}. Our contribution on
this axis is not a new inequality but a serving-system instantiation in
which every constant matches the implementation and the whole chain is
machine-checked in Lean/mathlib~\cite{mathlib} over a finite product
space, with no measure-theoretic axioms beyond the standard three.

\textbf{Runtime verification.} Monitoring executions against formal
specifications is a mature field~\cite{leucker-rv}; the usual setting is
discrete properties over event traces. Our monitors are numerical---error
radii, budgets, supermartingales---and the ``specification'' is a probability
statement whose side conditions (identity resolution, write-audit ordering,
per-event accounting) are enforced fail-closed and surfaced as all-zero gate
criteria. The concurrency guard that exposed a slot-namespace collision is,
in runtime-verification terms, a monitored invariant doing exactly what
monitors are for.

\textbf{Error propagation.} The composition rule \eqref{eq:compose} is affine
composition with a union bound, and the recurrent case is a discrete Gr\"onwall
inequality; we claim no novelty in either. The contribution is that heterogeneous
memory stages are expressed in one form, that the coefficients are computed from
live telemetry rather than assumed, and that the strength of the composed statement
is decided by the machine.

\section{Conclusion}
Heterogeneous attention memory breaks the implicit assumption behind existing
KV-cache certificates, including our own companion meter~\cite{witcert-companion}: that
there is one memory object to bound. We proposed typed
observability contracts---per-stage error bounds whose metric is part of the type,
composed only when metrics match, with proof obligations carried by name and tiers
decided by the machine---and instantiated them across latent caches, sparse
selectors, pooled hierarchies and recurrent states on production serving stacks.
The type discipline did real work: it rejected our own first composed bound, forced
two bridge theorems and one paired measurement into existence, and priced every
unprovable step into an explicit tier instead of a hedged sentence. On the systems
side, capture-time layer selection plus a fused meter make declared-subset
observation affordable under CUDA graphs, and an entry-level certificate ledger
with local fallback turns the per-stage bounds into a request-level
risk budget---sound at unknown length, with empirical objects excluded by
construction, executed so far at write-time policy granularity with read-side
shadow accounting. Applied to a served DeepSeek-V4 stack, the same machinery turned a silent
quality regression into a precise structural boundary---exact in the
eviction-free, identity-isolated regime, every observed failure in an
eviction or slot-reuse regime---with the microscopic mediator honestly recorded as open after
a machine-adjudicated campaign exhausted every instrumented channel; the
contract lesson, that a replaced memory representation redefines
coordinate, generation and materialization semantics, is the paper's most
transferable systems finding. What the calculus still cannot do is cross
layers: per-layer distributions are distinct objects, and no inter-layer
bridge is claimed. That boundary is not a rhetorical limitation; it is the
next theorem the type system is waiting for.

\appendix
\setcounter{figure}{0}\renewcommand{\thefigure}{A.\arabic{figure}}
\setcounter{table}{0}\renewcommand{\thetable}{A.\arabic{table}}
\section{The discrimination campaign, round by round}
\label{app:campaign}
This appendix preserves the full adjudication log summarized in
\S\ref{sec:served}. Every number is read mechanically from the
corresponding run artifact; nothing here is narrated from memory.

\begin{figure}[t]\centering
\includegraphics[width=0.92\linewidth]{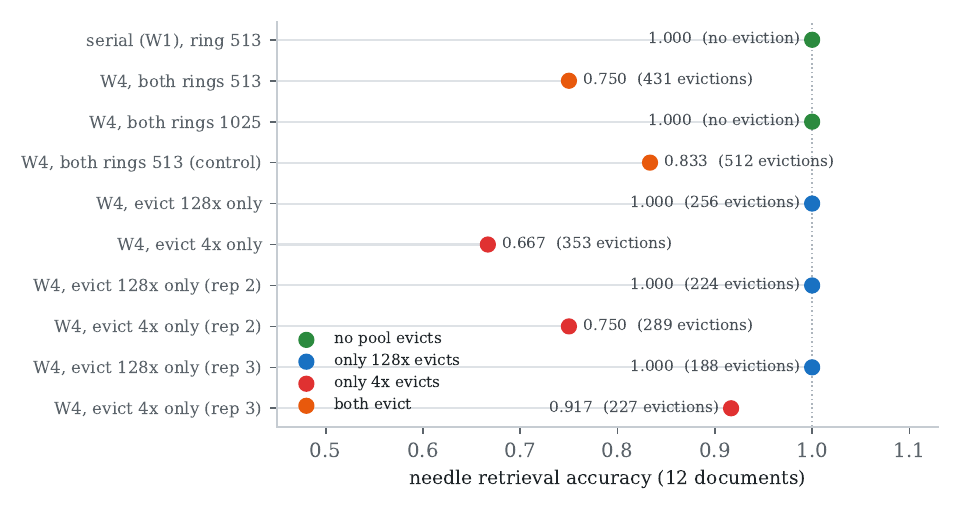}
\caption{\textbf{Nine-round controlled ablation.} Green: no
pool evicts (exact). Blue: only the $128\times$ ring evicts (exact in
all three repetitions). Red: only the $4\times$ ring evicts (degraded,
with run-to-run drift under identical configuration). Corruption tracks
$4\times$-ring eviction dose; every point is read from its run
artifact.}
\label{fig:localization}
\end{figure}

First, the ring
write/finalize/recycle chain passed a machine-adjudicated concurrent gate: a
per-\emph{(layer, logical-slot, write-epoch)} coverage ledger checked every
decoded slot reference against the generation that produced it, across both
arms and all eight ranks under full-ring pressure, with zero
out-of-generation reads---a diagnostic-chain claim
(ring$\to$finalize$\to$scratch-decode bookkeeping), not a claim about
the packed read path itself. Second, packed data is now \emph{consumed} in
serving rather than merely audited: a non-fused reference consumer
classifies every referenced slot as ring-resident, packed-final, or
fail-loud, and under an evict-invalidates-coverage semantics it
re-materialized $26{,}440$ rows from packed INT4 while both arms held exact
retrieval parity (the three-way identity
$\mathit{resident}+\mathit{packed}+\mathit{failed}+\mathit{skip}=\mathit{total}$
holding per rank, tolerance zero). Third, the $4\times$ pool---the
$32\times$ memory bulk deferred above---was replaced end-to-end with a
packed INT6 truth source under the same machinery: the formal gate (nine
criteria including a client-abort perturbation arm) passed with measured
HBM reduction of $2{,}108$\,MiB per GPU against a tensor-derived byte model
(all inputs read from live storage sizes, not derived estimates), and the
$128\times$ single-layer staging variant measured a median $316$\,MiB per
GPU net against its own byte model. Finally, the consumption chain was
brought inside CUDA graphs: with both graph modes enabled, the
packed-kernel wrapper recorded $5{,}576$ packed invocations ($2{,}720$
INT4 from the $128\times$ pool, $2{,}856$ INT6 from the $4\times$
pool) at exact retrieval parity on both arms---a capture-time and
eager-path call count, since graph replay executes recorded kernels
without re-entering host counters; that the replayed graphs contain the
packed branch follows from these being the branches taken at capture. The scope of that last claim is
deliberately narrow---six needle documents, one seed, one hardware
configuration, radix caching off, fixed ring sizes. Of the kernel
invocations, $272$ fell back to the native path; pointer-classified
attribution later showed every fallback call carries no compression
cache at all (it is not a compressed-pool read), so the fallback count
is a call-mix figure, not a coverage gap. The narrowness matters: a
twelve-document probe of the same configuration (no speculation)
subsequently scored $0.750$ with mid-depth misses that also recur under
speculative decoding---a systematic residual that the six-document
battery had been masking. A nine-round controlled-ablation chain then
localized the defect (Figure~\ref{fig:localization}), and we report the
chain because each round eliminated a hypothesis by construction.
 A
zero-eviction ring restores exactness under full four-way concurrency
($1.000$, against $0.833$ for the same round's small-ring control), and
the serial run that first suggested a concurrency skew turns out to have
evicted nothing at all---``serial is clean'' was a confound, and the
invariant across all rounds is that corruption appears exactly when
pages of in-flight requests are evicted. A per-pool dissociation pins
the pool: evicting only the $128\times$ ring is exact (reproduced three
times), evicting only the $4\times$ ring degrades to $0.667$. The write
side is exonerated by row-level forensics: $16{,}942{,}020$
bake-versus-store checks across both pools found zero ownership
mismatches, and an in-graph counter of finalizations touching freed
pages read zero. The consumer was then found by a two-step probe.
Coordinate-domain instrumentation showed that the
compressed-context (``extra'') reads address the ring-registered buffers in
cache-wide coordinates---observed indices reach $66\times$ the
$4\times$ ring's row count---and a contribution ablation that zeroes
this path's top-$k$ lengths collapses retrieval to $0.000$ against a
$0.667$ in-band control (the arms differ in downstream eviction counts,
so this is a necessary-channel localization, not a dose-matched effect
size): the reads are \emph{essential}, and their data
is whatever eviction leaves behind, which is why corruption tracks
eviction dose rather than any timing discipline (two fence designs and
an aged-reuse quarantine, all mechanically effective, all failed to
help). The in-graph claim above is therefore a six-document,
eviction-free-regime claim; a ring sized to the live working set is the
demonstrated clean configuration. A prefill-side materialization
adapter was then trialled by forcing the sparse-prefill branch, whose
dequant anchor carries the translation-and-materialization shims: the
corrected reads executed at scale ($8{,}023{,}680$ row translations,
zero orphans) yet quality did not recover under eviction---correct
addressing of the \emph{prefill} phase is not the harmful mediator.
The consumer was finally pinned on the decode side by closing the state
space. A content oracle over the $4\times$ pool's written entries (an
$L^1$-norm ordered-multiset metric, judged only against a
same-configuration repeat pair's calibrated noise floor) found the
corrupt and healthy arms' persistent contents statistically identical
(median relative difference $0.0026$ against a $0.0021$ floor); the
same instrument over the sliding-window pool---whose rows are a
function of the prefill residual stream---likewise read at its floor
(median $0.0016$ against a $0.0010$ floor),
so the entire prefill computation is arm-invariant and the harm is
confined to decode-time consumption. The read-side coordinate exposure is
then arithmetic: compressed-context reads index the backing stores by
\emph{logical} slot (the read-side translation counter reads zero
across all rounds, and the observed index bound exceeds the small
ring's physical capacity), while ring writes land at \emph{translated
physical} slots. An intervention isolates an effective lever: folding
reads of non-resident (evicted) and out-of-range entries to the
kernel's skip sentinel---honest truncation rather than silent
consumption---restores the eviction microcosm from $0.500$ to $1.000$
against a knob-isolated control with matched eviction and call counts,
the cleanest premise set in the adjudication chain. Scope is stated
with the full honesty the campaign earned. The truncation lever is
timing-sensitive: when eviction overlaps the prefill window the same
truncation costs the needle's context and the microcosm fails at
$0.500$, so this is a mitigation, not a repair. At the twelve-document
concurrent scale a four-pair dose-stratified design plus a
control-variance baseline shows eviction dose itself is
scheduling-stochastic (a same-configuration control pair spans
$0.667$--$1.000$), and at matched dose the truncated and untreated
arms sit on one dose--response curve: truncation and corruption are
quality-equivalent harms on this retrieval task. Deepest of all, after
nine further instrumented rounds---selection streams judged against a
repeat-pair noise floor, served-content ledgers, archive-completeness
audits at eviction time, key-space provenance---every measured
quantity is arm-equal to its calibrated floor, the archive is complete
and correctly keyed, and the quality split persists; the microscopic
final mile of the harm mechanism is therefore recorded as an open
question rather than narrated beyond the evidence. \emph{Postscript
(the rounds after the digest was frozen).} Four further adjudications
sharpened the exclusion list. A repeat of the intervention pair with
early-landing evictions failed at $0.500$ in both arms, establishing the
truncation lever's timing sensitivity. A key-space provenance oracle
proved the packed archive logically keyed (readback matches the archive
at INT6 quantization scale, $0.092$ maximum relative difference, against
$1.33$ for the physical-key hypothesis). An eviction-time
archive-completeness audit read $0.0216$ mean relative difference
($0.028$ max, zero unwritten rows) between the evicted pages' archive
and the ring's final values---the archive is complete. Finally, a
materialized reference reader (the instrument-verified dequantization
path plus reference attention mathematics, $1{,}430$ calls in a matched
arm) reproduced the control's $0.500$ exactly, exonerating the
production kernel's inline unpack and the reader mathematics; selection
streams judged against a repeat-pair noise floor showed top-$k$
scrambling is run-to-run nondeterminism, quality-irrelevant in clean
pairs. Every one of these instruments narrowed the mediator's possible
location; none found it. The claim in the main text is calibrated
accordingly.

\section{Per-path evidence and measurement regimes}
\label{app:paths}
CUDA-graph serving: in-graph packed consumption is exact at the
six-document eviction-free regime ($1.000/1.000$); the twelve-document
concurrent probe sits at $0.75$ under the eviction mechanism of
\S\ref{sec:served}, so the path's claim is regime-limited rather than
unconditional. Repeat-run answer stability passes ($1.000/1.000$ across a
same-config serial repeat pair; text-level determinism is explicitly not
claimed, as the upstream deterministic-inference mode does not support
this attention backend). Radix prefix caching fails reproducibly with the
method---the second round after the read path was fully wired, excluding
an instrumentation confound; lifting this requires
prefix-reuse--aware generation semantics. Speculative decoding: its
translation chain closed first (the fail-loud counters that once fired
stay zero); an earlier cross-commit comparison suggesting nondeterminism
was methodologically invalid, and a same-source rerun showed the misses
recur without speculation; the two-seed rerun at the working-set-ring
regime is zero-eviction, machine-annotated in the artifact---each round
its own dual-arm control.

\section{Always-on probes under CUDA graphs: the engineering account}
\label{app:overhead}
Under a
graph-enabled run, the host-side call counter grew by $168$, exactly the
prefill calls ($6$ requests $\times$ $28$ layers), while the device-side
accumulators grew by $43176$, exactly the $43008$ decode replays plus those
same prefill calls; the counts were unchanged after an idle interval, ruling out a
lagging flush. Since replay runs no Python, that growth can only come from probe
operations captured into the graph. Four changes were necessary to get there, each
worth stating because each is a silent-corruption trap rather than a crash:
accumulators \emph{including counters} must be device-resident tensors, since a
host-side count would freeze at capture while the sums kept growing, inflating every
mean; the sampling path must avoid RNG, which invalidates capture; flushing must be
deferred outside the capture window, since reading accumulators back is a device
synchronisation; and persistent accumulators must not be allocated during capture,
because serving captures one graph per batch size and an accumulator allocated inside
one graph's memory pool is invalid from another. The packed-consumption
runs of \S\ref{sec:served} rediscovered the same trap classes on the write
side---a call-site argument evaluated before its guard, a lazily built
device constant, a Python boolean scalar assigned into a device tensor, a
data-dependent boolean-mask index---each invalidating capture rather than
crashing, and each now a regression test.

Running under graphs is not the same as running always-on under graphs, and getting
there took two changes rather than one. At full rate the naive cost is prohibitive: on
a single-GPU Qwen2.5-7B serving stack with graphs enabled, throughput falls
$-52.8\%$ and TTFT P99 rises $+138.6\%$ against an identical probe-off
baseline. The obvious remedy---sample one call in $N$, which is what makes the probe
affordable with graphs disabled---does not transfer, because the sampling test is a
Python branch and Python runs only at capture. Its value at that instant decides
whether a given graph contains probe operations at all, so $N$ stops meaning ``one
call in $N$'' and starts meaning ``these graphs, never those''. Measured at $N=64$
under graphs, the accumulators grew by $0$ across an entire workload while the
overhead was correspondingly nil---a probe that is not in the graph is indeed free.
Worse, that run \emph{passed} our acceptance gate, which saw all $28$ layers covered
by samples left over from start-up, before graph capture began. Coverage over the
run's lifetime is not liveness during the run. We therefore added a device-side
execution counter, incremented in place and hence captured into the graph, and a gate
check requiring it to exceed the host-side call count; the element counts already
collected cannot serve this purpose, since they cannot distinguish one call sampling
many rows from many replays sampling one row each.

The first change is to make the sampling decision depend only on quantities that are
constant across replays. A layer index is such a quantity, and selecting instrumented
layers by index behaves as intended: at every stride we tried, replay accumulation is
present and the observed layers match the declared set exactly. It is not sufficient
on its own. Cost tracks the number of instrumented call sites and the number of
kernels each site issues, and the two scale close to linearly and multiply: halving
the quantiser work at a fixed two layers cuts the throughput cost by $1.72\times$,
and going from two instrumented layers to one at a fixed single quantiser cuts it by a
further $1.97\times$, reaching $-2.26\%$ throughput and $+1.32\%$
TTFT P99---still outside the noise floor. Sampling less cannot remove a
per-execution constant, so the
second change attacks the constant itself. We fuse the witness meter into a single
Triton kernel that loads the sampled rows once, performs the per-band quantisation and
both norms in registers, and folds the result into the accumulators, replacing several
dozen kernel launches with one. It agrees with the reference implementation to a
maximum relative error of $9.3e-07$ on every statistic, with violation counts
and sample counts exactly equal---the violation count is a soundness self-check and is
not allowed to be approximate---and a single invocation costs $12.3$~us, or
$24.2\times$ less than the unfused single-quantiser path. At equal, full
$28$-layer coverage the fused meter is $5.2\times$ cheaper than the
unfused one.

With both changes, one configuration meets all three conditions we set in advance:
instrumenting one layer in twenty-eight with the fused meter costs $-0.94\%$
throughput and $-0.14\%$ TTFT P99, both inside the per-concurrency noise
floor, while replay accumulation is present and the observed coverage matches the
declared single layer. We state the scope precisely: what is
demonstrated is always-on observation of \emph{a declared subset}, not of
every layer.

\section{Certificate-guided low-bit format exploration}
\label{app:casestudy}
As a boundary-discovery exercise---not a completed compression system---we
used the instrument to drive an aggressive low-bit format study on V4-Flash's
compressed pools. The witness ladder selected block-Hadamard INT6 for the
memory-dominant $4\times$ pool and INT4 for the $128\times$ pool on hidden
requests (naive INT4 rejected: $2\%$ coverage, $15\times$ the INT8 control's
score-shift proxy), a fused unpack kernel measured $3\times$ \emph{faster}
than the production FP8 dequantizer (bit-exact, tensor-core inverse
Hadamard), and a real $360$\,B/token packed pool---$36.5\to 22.5$\,MiB for
$64$Ki tokens as a standalone allocation, a format-level figure we do not
conflate with serving-process memory---ran in serving with
$12{,}960$ read-back checks, zero unwritten slots, and $10.2\%$ mean relative
error. The instrument then falsified two of our own design assumptions: entry
slots evolve progressively, so write-time packing reads a non-final state
(changing only the packing instant moved the discrepancy $18\times$); and the
apparent Python read path is not the production consumer---decode reads FP8
pages inside the FlashMLA kernel, so three instrumented probes of the swap
path never fired. We therefore claim format, kernel, pool, and write-side
lifecycle as validated, and explicitly do \emph{not} claim served INT4
consumption, HBM reduction in the serving process, throughput or quality
effects: those require a packed-input FlashMLA variant, which the instrument
has now precisely scoped. The residual $\mathrm{rel}_{\max}\approx 4.9$
traces to re-reads of still-evolving slots, not quantization failure
(the near-zero-norm hypothesis was tested and rejected).

\section{The staged-ring discharge ledger}
\label{app:discharge}
The claims withheld above can now be partially discharged. Served consumption
holds: with invalidation in place the packed pools serve the retrieval task at
parity ($1.000$, $20/20$, zero request errors). Structural pool replacement
holds: swapping the $128\times$ pool to the packed object at the canonical
address (twenty layers remapped, installation attested per rank) is
quality-neutral (acc $1.000$, decode-side fidelity unchanged). Physical
memory reduction holds at mechanism level: shrinking the FP8 staging pool to
a $257$-page window ring freed $2{,}988$\,MiB of measured device memory
across eight GPUs against an accounting estimate of $3{,}040$\,MiB---an
aggregate agreement of $1.7\%$, though the per-GPU deltas are non-uniform
($566$ on rank~0, $346$ elsewhere), so we claim aggregate agreement only.
We state the measurement regime plainly: the $128\times$
pool is ${\sim}0.3\%$ of process memory, and the $32\times$-larger $4\times$
pool, where the same construction must be repeated for material savings,
remains future work. One claim is explicitly \emph{not} discharged: the
ring-staged configuration initially degraded retrieval to $0.500$ vs.\
$1.000$ while every fail-loud surface stayed green---and the artifact
carried no internal counters (an evidence-chain gap we record rather than
paper over), so only the HTTP-level zero-error figure is attested. A
scoped intervention later identified an addressing defect as a major cause
(the production decode kernel indexing the $257$-page ring with
$20{,}413$-page logical slot ids---an out-of-bounds silent read); at that
stage, whether the write/finalize/recycle chain was itself loss-free
remained open pending a port-separated re-run with poisoned-scratch
instrumentation, and the stage stood as memory-mechanism evidence only.

\section{Ledger enforcement mechanics}
\label{app:ledgermech}
Probabilistic budget is spent with \emph{pre-authorization} semantics: before any
random bits are drawn, a prior radius $u_e$ (a mean term plus a
$\sqrt{2\Sigma s_c^2\ln(n/\delta_i)}$ tail computed from the stored exponents
alone) authorizes an entry only if $u_e$ clears the threshold; unauthorized
entries fall back without drawing. Each $\delta_i$ from the telescoping sequence
is thereby bound to a concrete event---some authorized entry's realized witness
exceeding its pre-declared radius---and a write-side audit counts realized
violations: $0$ of $147{,}208$ authorized entries violated their radius,
validating the conditional-rate semantics at runtime
($\Sigma\delta=0.0099997<\delta_{\mathrm{req}}$ over $33{,}407$ events,
with the moment-condition assumption logged with per-event counts). The price of
authorizing before the draw is quantified rather than hidden: the prior radius
authorizes $6.2\times$ fewer entries than post-hoc deterministic gating
accepts ($7.8\%$ vs.\ $48.0\%$).

Four runtime
invariants make this shared-read semantics fail-closed rather than
observational: the write-time audit completes inside the same hook call,
before any shared read; every slot carries an explicit state
(unwritten / masked / kept-exact / restored-exact); a violating entry has its
original bytes restored on the spot, so the served value respects the
declared radius \emph{deterministically}---``zero violations'' is enforced,
not merely observed; and an immutable write event spends its $\delta$
exactly once however many requests later read it, which is the content of a
machine-checked lemma (\texttt{shared\_read\_sound}: the union bound ranges
over write events, never over readers). Identity resolution itself is
fail-closed: a batch whose per-entry request attribution cannot be
established keeps exact bytes, spends no probabilistic budget, and increments
a counter that the acceptance gate requires to be zero---the gate is a
machine-decidable conjunction of twelve all-zero criteria (including
missing-data checks: all eight replica snapshots present, sentinel factors
nonempty) carried by the adjudication script's exit code. The restore step
upgrades the online semantics, and we state it precisely to avoid a reading
we do not claim: admission is \emph{probabilistic}, enforcement is
\emph{deterministic}. The request-level $\delta$ bounds the probability of
\emph{unexpected restoration events}---an authorized entry's realized
witness exceeding its pre-declared radius, forcing the audit to restore
exact bytes---not the probability of serving an out-of-bound value: the
served value respects its declared radius always, by the write-audit and
restore invariants above. Shared readers inherit that deterministic
guarantee and never multiply the originating write event's risk.

Request identity is a monotone logical UID that is never
reused: the boundary is detected \emph{on the write path itself}, at the
non-extend-to-extend transition that opens every request's prefill, so it
precedes that request's first write and account boundaries coincide with true
request boundaries (a fallback on sequence-length drops covers read-only
deployments; the reuse-with-longer-prompt case the fallback alone would miss
is pinned by a unit test). Each request carries an \emph{independent}
telescoping budget: the eight user-request accounts---reported separately
from one server-warmup account---each stay under budget with maximum spend
$\Sigma\delta = 0.009982 \le \delta_{\mathrm{req}} = 0.01$, and
per-request event counters restart from $\delta_1 = \delta_{\mathrm{req}}/2$
(the earlier implementation let one global counter starve later requests).
The stochastic-rounding seed is the quadruple (server nonce, request UID,
layer, event counter): distinct requests never share a random stream, so
conditional randomness holds under adaptive traffic, while the quadruple is
rank-invariant---the eight tensor-parallel replicas realize the \emph{same}
draw, agree field-for-field in their accounts, and $\delta$ is spent once
per logical event, an empirically checked coupling rather than an assumption.

\section{The generation-staleness discrimination}
\label{app:staleness}
This appendix gives the full evidence chain behind the boundary closure
stated in \S\ref{sec:generation}. A long-context retrieval task
(needle-in-haystack, exact match, matched five-arm control) first exposed a
$95$-point collapse that radix caching had been masking: the packed arm scored
$0.800$ with the cache on---a cache-hit figure, not a quality figure---and
$0.050$ with it off, while the FP8 baseline held $1.000$ on both sides. A
freshness discriminator (identical arms differing only in a
re-pack-on-every-read diagnostic, both write- and read-path execution
counters required to be nonzero) recovered $0.050$ vs.\ $1.000$: generation
consistency, not quantization, was the active hypothesis. Offline round-trips
of the production pack path on real entries pinned the quantizer's intrinsic
relative error at $0.0245$ (kernel, pool path, and overwrite all agreeing;
maximum $0.0304$; the RoPE segment exactly zero), so the $\mathrm{rel}>1$
values observed in serving were not achievable by the quantizer on this data
at all---they were reads of a \emph{previous generation}. The mechanism: slot
reuse by the allocator with no invalidation call anywhere in the codebase, so
a ``written'' flag conflated \emph{has ever been packed} with \emph{current
generation}. Unconditional write-site invalidation (a boolean scatter plus a
generation increment, placed before any sampling gate) restored task accuracy,
and a second causal confirmation with the diagnostic disabled---arms differing
only in the invalidation switch, with two-sided evidence (the fix arm's
invalidation counter positive, the control arm's exactly zero)---returned
\textsc{invalidation\_confirmed} on six machine-checked criteria: $1.000$
vs.\ $0.200$ under $4$-way concurrency.

\section{The repaired chain, fully accounted}
\label{app:composition}
This appendix accounts, term by term, for how both scope restrictions were
discharged by measurement. The probe now witnesses the compressed-page entries the selector chooses
(their witness is \emph{larger} than the window's, max $1.040$ against
$0.700$, so treating pages as exact had been the optimistic side), and on the
$21$ sparse-selection layers where every term is witnessed the full chain gives
a worst-case per-layer TV \emph{bound} of $1.473$ (median; $4/21$
layers below $1.0$) and a mean-coefficient \emph{diagnostic reading} of
$0.992$ (median)---the diagnostic has no bound semantics. For the selection term, the missing
selector-to-attention bridge is replaced by a paired measurement on identical
candidate sets: the ratio of attention-mass omitted to selector-mass omitted is
$0.87$ (median; tail to $2.18$, and we convert with the tail value),
so $b_S$ crosses into attention units as $0.00444$ at tier
\emph{empirical}---and the composed tier drops to empirical with it, automatically.
The same measurement separates two quantities our first analysis conflated: the
top-$512$ truncation itself leaves $0.379$ (median) of the hypothetical
full-compressed attention mass unattended, but that is the architecture's designed
sparsity, not an error---the model's ideal reference includes the truncation, and
only perturbation-induced \emph{changes} of the selection belong in the error
chain. These numbers
are larger than the retracted ones because they are denominated in a real metric and
pay the full price of soundness: Cauchy--Schwarz assumes the perturbation aligns
with the query, and the witness overstates the residual by construction. A bound
that got smaller by adding incompatible units was not a bound. The self-checks
remain $0$ throughout, and the sensitivity ordering says where tightening pays: the
witness $W$ first, the alignment worst case second; $b_S=0.00203$, in its own
units, is not the bottleneck.
\section{Machine-checked theorem statements}
\label{sec:theorems}

Compiled, not transcribed. Every theorem name cited in the body (\texttt{ledger\_sound},
\texttt{telescope\_sum}, \texttt{shared\_read\_sound}, the bridge and
e-process lemmas, \ldots) resolves to a statement below. This section is
\emph{compiled} from the Lean development shipped in the released repository (\texttt{witprobe-atten\-tion-memory}): \texttt{formal/Export.lean} walks the
environment, prints each theorem's \emph{elaborated} type with Lean's own
pretty-printer, and lists the axioms each proof actually depends on via
\texttt{collectAxioms}; non-ASCII symbols are mechanically transliterated for
the pdflatex toolchain. Nothing here is transcribed by hand, so a statement
that is not proved cannot appear. The contract calculus proves 14 \emph{core} theorems---the results cited in the body---plus 143 supporting lemmas listed compactly so the axiom audit covers the whole
development; every proof depends only on \texttt{propext},
\texttt{Classical.choice} and \texttt{Quot.sound}, and none on \texttt{sorryAx}.
The whole development re-verifies independently: clone
\url{https://github.com/metask-ai/witprobe-attention-memory} and run
\texttt{cd formal \&\& bash check\_all.sh}, which rebuilds every proof and re-runs the
axiom audit from scratch; the \texttt{file:line} references below resolve in that repository.

\paragraph{conditional (adaptive) union bound over first-failure events}
{\small\raggedright \emph{Name:} \texttt{WitCert.Calculus.Budget.conditional\_union\_le}; \emph{source:} \texttt{formal/WitCert/Contract.lean:143}; \emph{axioms:} \texttt{propext, Classical.choice, Quot.sound}.\par}
{\scriptsize\begin{verbatim}
forall  {Omega : Type u_1} [inst : MeasurableSpace Omega] (mu : MeasureTheory.Measure Omega) (E
    : Nat -> Set Omega) (delta : Nat -> ENNReal),
  (forall  (i : Nat), mu (E i \ Union  j  in  Finset.range i, E j) <= delta i) ->
    forall  (n : Nat), mu (Union  i  in  Finset.range n, E i) <= Sum  i  in  Finset.range n,
    delta i
\end{verbatim}}

\paragraph{coverage confidence: (1-p)\textasciicircum{}n <= e\textasciicircum{}\{-np\}; zero observed violations is not zero rate}
{\small\raggedright \emph{Name:} \texttt{WitCert.Calculus.Ledger.coverage\_confidence}; \emph{source:} \texttt{formal/WitCert/Ledger.lean:89}; \emph{axioms:} \texttt{propext, Classical.choice, Quot.sound}.\par}
{\scriptsize\begin{verbatim}
forall  {p : R}, 0 <= p -> p <= 1 -> forall  (n : Nat), (1 - p) ^ n <= Real.exp (-(n * p))
\end{verbatim}}

\paragraph{request-level ledger: conditional union bound + budget sequence, sound at every length}
{\small\raggedright \emph{Name:} \texttt{WitCert.Calculus.Ledger.ledger\_sound}; \emph{source:} \texttt{formal/WitCert/Ledger.lean:66}; \emph{axioms:} \texttt{propext, Classical.choice, Quot.sound}.\par}
{\scriptsize\begin{verbatim}
forall  {Omega : Type u_1} [inst : MeasurableSpace Omega] (mu : MeasureTheory.Measure Omega) (E
    : Nat -> Set Omega) (deltaseq : Nat -> ENNReal)
  (deltareq : ENNReal),
  (forall  (i : Nat), mu (E i \ Union  j  in  Finset.range i, E j) <= deltaseq i) ->
    (forall  (n : Nat), Sum  i  in  Finset.range n, deltaseq i <= deltareq) -> forall  (n :
    Nat), mu (Union  i  in  Finset.range n, E i) <= deltareq
\end{verbatim}}

\paragraph{refinement interface: any CertifiedEventStream satisfies the ledger guarantee}
{\small\raggedright \emph{Name:} \texttt{WitCert.Calculus.Ledger.stream\_sound}; \emph{source:} \texttt{formal/WitCert/Ledger.lean:204}; \emph{axioms:} \texttt{propext, Classical.choice, Quot.sound}.\par}
{\scriptsize\begin{verbatim}
forall  {Omega : Type u_2} [inst : MeasurableSpace Omega] {mu : MeasureTheory.Measure Omega}
  (s : WitCert.Calculus.Ledger.CertifiedEventStream Omega mu) (n : Nat), mu (Union  i  in
    Finset.range n, s.E i) <= s.budget
\end{verbatim}}

\paragraph{telescoping budget weights 1/((i+1)(i+2)): unknown-length budget sums to 1-1/(n+1)}
{\small\raggedright \emph{Name:} \texttt{WitCert.Calculus.Ledger.telescope\_sum}; \emph{source:} \texttt{formal/WitCert/Ledger.lean:37}; \emph{axioms:} \texttt{propext, Classical.choice, Quot.sound}.\par}
{\scriptsize\begin{verbatim}
forall  (n : Nat), Sum  i  in  Finset.range n, WitCert.Calculus.Ledger.w i = 1 - 1 / (n + 1)
\end{verbatim}}

\paragraph{two-layer budget: deterministic events at delta=0, probabilistic on telescoping counters}
{\small\raggedright \emph{Name:} \texttt{WitCert.Calculus.Ledger.two\_layer\_budget\_le}; \emph{source:} \texttt{formal/WitCert/Ledger.lean:144}; \emph{axioms:} \texttt{propext, Classical.choice, Quot.sound}.\par}
{\scriptsize\begin{verbatim}
forall  (delta : R),
  0 <= delta ->
    forall  (f : Nat -> Option Nat),
      (forall  (i k : Nat), f i = some k -> k = WitCert.Calculus.Ledger.probCount f i) ->
        forall  (n : Nat),
          (Sum  i  in  Finset.range n,
              match f i with
              | none => 0
              | some k => ENNReal.ofReal (delta * WitCert.Calculus.Ledger.w k)) <=
            ENNReal.ofReal delta
\end{verbatim}}

\paragraph{unknown-length budgeting costs at most sqrt(2) in sub-Gaussian radius vs uniform split}
{\small\raggedright \emph{Name:} \texttt{WitCert.Calculus.Ledger.unknown\_length\_price}; \emph{source:} \texttt{formal/WitCert/Ledger.lean:112}; \emph{axioms:} \texttt{propext, Classical.choice, Quot.sound}.\par}
{\scriptsize\begin{verbatim}
forall  {delta : R}, 0 < delta -> delta <= 1 / 2 -> forall  {N i : Nat}, 1 <= i -> i <= N ->
    Real.log (i * (i + 1) / delta) <= 2 * Real.log (N / delta)
\end{verbatim}}

\paragraph{e-process: product of conditionally mean-<=1 factors gives anytime-valid risk <= delta}
{\small\raggedright \emph{Name:} \texttt{WitCert.Calculus.Ville.eprocess\_ville}; \emph{source:} \texttt{formal/WitCert/Ville.lean:127}; \emph{axioms:} \texttt{propext, Quot.sound, Classical.choice}.\par}
{\scriptsize\begin{verbatim}
forall  {sigma : Type u_1} [inst : Fintype sigma] (D : WitCert.Calculus.Ville.Draw sigma) (g :
    List sigma -> sigma -> R) {delta : R},
  0 < delta ->
    delta <= 1 ->
      (forall  (h : List sigma) (x : sigma), 0 <= g h x) ->
        (forall  (h : List sigma), Sum  x : sigma, D.p x * g h x <= 1) ->
          forall  (T : Nat), WitCert.Calculus.Ville.hitProb D
    (WitCert.Calculus.Ville.prodProcess g) (1 / delta) T [] <= delta
\end{verbatim}}

\paragraph{finite-Omega Ville inequality: P(sup M >= c) <= M0/c, elementary induction, no measure theory}
{\small\raggedright \emph{Name:} \texttt{WitCert.Calculus.Ville.ville}; \emph{source:} \texttt{formal/WitCert/Ville.lean:65}; \emph{axioms:} \texttt{propext, Quot.sound, Classical.choice}.\par}
{\scriptsize\begin{verbatim}
forall  {sigma : Type u_1} [inst : Fintype sigma] (D : WitCert.Calculus.Ville.Draw sigma) (M :
    List sigma -> R) {c : R},
  0 < c ->
    (forall  (h : List sigma), 0 <= M h) ->
      (forall  (h : List sigma), Sum  x : sigma, D.p x * M (x :: h) <= M h) ->
        forall  (T : Nat) (h : List sigma), WitCert.Calculus.Ville.hitProb D M c T h <= M h / c
\end{verbatim}}

\paragraph{bridge contract semantics: cross-metric composition needs a proved bridge}
{\small\raggedright \emph{Name:} \texttt{WitCert.Calculus.bridge\_sound}; \emph{source:} \texttt{formal/WitCert/Contract.lean:118}; \emph{axioms:} \texttt{propext, Classical.choice, Quot.sound}.\par}
{\scriptsize\begin{verbatim}
forall  {alpha : Type} {m m' : WitCert.Calculus.ErrMetric alpha} (B : WitCert.Calculus.Bridge m
    m') (x x' : alpha),
  m'.d x x' <= B.toContract.a * m.d x x' + B.toContract.b
\end{verbatim}}

\paragraph{typed contract composition: metrics must match by construction}
{\small\raggedright \emph{Name:} \texttt{WitCert.Calculus.comp\_sound}; \emph{source:} \texttt{formal/WitCert/Contract.lean:93}; \emph{axioms:} \texttt{propext, Classical.choice, Quot.sound}.\par}
{\scriptsize\begin{verbatim}
forall  {alpha beta ? : Type} {malpha : WitCert.Calculus.ErrMetric alpha} {mbeta :
    WitCert.Calculus.ErrMetric beta}
  {m? : WitCert.Calculus.ErrMetric ?} (C2 : WitCert.Calculus.Contract mbeta m?) (C1 :
    WitCert.Calculus.Contract malpha mbeta)
  (x x' : alpha), m?.d (C2.exact (C1.exact x)) (C2.approx (C1.approx x')) <= C2.a * C1.a *
    malpha.d x x' + (C2.a * C1.b + C2.b)
\end{verbatim}}

\paragraph{witness-to-score bridge: |Delta(scale*q*k)| <= scale*||q||*||Deltak|| (Cauchy-Schwarz)}
{\small\raggedright \emph{Name:} \texttt{WitCert.Calculus.score\_bridge}; \emph{source:} \texttt{formal/WitCert/Bridges.lean:36}; \emph{axioms:} \texttt{propext, Classical.choice, Quot.sound}.\par}
{\scriptsize\begin{verbatim}
forall  {d : Nat} (q k k' : Fin d -> R) (scale : R),
  0 <= scale ->
    |scale * Sum  i : Fin d, q i * k i - scale * Sum  i : Fin d, q i * k' i| <=
      scale * sqrt (Sum  i : Fin d, q i ^ 2) * sqrt (Sum  i : Fin d, (k i - k' i) ^ 2)
\end{verbatim}}

\paragraph{affine form of the score-to-TV bridge: TV <= e\textasciicircum{}\{2eps0\}*eps for eps <= eps0}
{\small\raggedright \emph{Name:} \texttt{WitCert.Calculus.softmax\_tv\_affine}; \emph{source:} \texttt{formal/WitCert/Bridges.lean:160}; \emph{axioms:} \texttt{propext, Quot.sound, Classical.choice}.\par}
{\scriptsize\begin{verbatim}
forall  {iota : Type u_1} [inst : Fintype iota] [inst_1 : Nonempty iota] (s s' : iota -> R)
    (eps eps0 : R),
  0 <= eps ->
    eps <= eps0 ->
      (forall  (t : iota), |s t - s' t| <= eps) ->
        WitCert.TV (WitCert.Calculus.softmax s') (WitCert.Calculus.softmax s) <= Real.exp (2 *
    eps0) * eps
\end{verbatim}}

\paragraph{score-to-TV bridge: ||Deltas||inf <= eps ==> TV <= 1/2(e\textasciicircum{}\{2eps\}-1), instantiating paper 1's tv\_le\_eform}
{\small\raggedright \emph{Name:} \texttt{WitCert.Calculus.softmax\_tv\_bridge}; \emph{source:} \texttt{formal/WitCert/Bridges.lean:93}; \emph{axioms:} \texttt{propext, Quot.sound, Classical.choice}.\par}
{\scriptsize\begin{verbatim}
forall  {iota : Type u_1} [inst : Fintype iota] [inst_1 : Nonempty iota] (s s' : iota -> R)
    (eps : R),
  0 <= eps ->
    (forall  (t : iota), |s t - s' t| <= eps) ->
      WitCert.TV (WitCert.Calculus.softmax s') (WitCert.Calculus.softmax s) <= 1 / 2 *
    (Real.exp eps ^ 2 - 1)
\end{verbatim}}

\paragraph{Supporting lemmas (axiom-audit completeness only)}
{\small\raggedright \texttt{Apriori.TV\_le\_one} (\texttt{WitCert/Apriori.lean:273}); \texttt{Apriori.TV\_nonneg} (\texttt{WitCert/Apriori.lean:268}); \texttt{Apriori.bdd\_diff\_of\_lipschitz} (\texttt{WitCert/Apriori.lean:661}); \texttt{Apriori.bdd\_diff\_of\_lipschitz\_on} (\texttt{WitCert/Apriori.lean:745}); \texttt{Apriori.bernoulli\_mgf\_le} (\texttt{WitCert/Apriori.lean:146}); \texttt{Apriori.betting\_factor\_eprocess} (\texttt{WitCert/Apriori.lean:803}); \texttt{Apriori.betting\_ucb\_anytime} (\texttt{WitCert/Apriori.lean:832}); \texttt{Apriori.condE\_eq\_prod\_sum} (\texttt{WitCert/Apriori.lean:470}); \texttt{Apriori.coord\_mgf\_le} (\texttt{WitCert/Apriori.lean:71}); \texttt{Apriori.doob\_mgf\_le} (\texttt{WitCert/Apriori.lean:257}); \texttt{Apriori.doob\_mgf\_le\_biased} (\texttt{WitCert/Apriori.lean:241}); \texttt{Apriori.doob\_mgf\_prod\_sum} (\texttt{WitCert/Apriori.lean:510}); \texttt{Apriori.eprocess\_factor\_le\_one\_of\_expected} (\texttt{WitCert/Apriori.lean:176}); \texttt{Apriori.iterComp\_mem} (\texttt{WitCert/Apriori.lean:757}); \texttt{Apriori.lip\_comp} (\texttt{WitCert/Apriori.lean:672}); \texttt{Apriori.lip\_iterComp} (\texttt{WitCert/Apriori.lean:713}); \texttt{Apriori.lip\_iterComp\_on} (\texttt{WitCert/Apriori.lean:770}); \texttt{Apriori.lip\_residual} (\texttt{WitCert/Apriori.lean:692}); \texttt{Apriori.mgf\_tensorize} (\texttt{WitCert/Apriori.lean:40}); \texttt{Apriori.mgf\_tensorize\_le} (\texttt{WitCert/Apriori.lean:53}); \texttt{Apriori.prod\_measure\_total} (\texttt{WitCert/Apriori.lean:28}); \texttt{Apriori.request\_tail\_of\_expected\_loss} (\texttt{WitCert/Apriori.lean:204}); \texttt{Apriori.request\_tail\_of\_served\_tv} (\texttt{WitCert/Apriori.lean:311}); \texttt{Apriori.request\_tail\_of\_served\_tv\_doob\_massweighted} (\texttt{WitCert/Apriori.lean:611}); \texttt{Apriori.request\_tail\_of\_served\_tv\_general} (\texttt{WitCert/Apriori.lean:376}); \texttt{Apriori.request\_tail\_of\_served\_tv\_massweighted} (\texttt{WitCert/Apriori.lean:420}); \texttt{Apriori.request\_tail\_of\_served\_tv\_topk} (\texttt{WitCert/Apriori.lean:531}); \texttt{Apriori.served\_tv\_mean\_le\_cum\_C} (\texttt{WitCert/Apriori.lean:105}); \texttt{BC\_ge\_subgaussian} (\texttt{WitCert/Bridges.lean:568}); \texttt{BC\_le\_one} (\texttt{WitCert/Bridges.lean:522}); \texttt{BatchEdit.applyAll\_isSome\_iff\_allHit} (\texttt{WitCert/BatchEdit.lean:59}); \texttt{BatchEdit.changed\_but\_not\_all\_applied} (\texttt{WitCert/BatchEdit.lean:81}); \texttt{BatchEdit.one\_hit\_masks\_many\_misses} (\texttt{WitCert/BatchEdit.lean:91}); \texttt{BatchEdit.overlap\_makes\_order\_matter} (\texttt{WitCert/BatchEdit.lean:101}); \texttt{Budget.conditional\_union\_le\_of\_marginal} (\texttt{WitCert/Contract.lean:173}); \texttt{Conformal.average\_does\_not\_give\_per\_history} (\texttt{WitCert/Conformal.lean:206}); \texttt{Conformal.conformal\_exceedance\_le} (\texttt{WitCert/Conformal.lean:97}); \texttt{Conformal.conformal\_risk\_union} (\texttt{WitCert/Conformal.lean:130}); \texttt{Conformal.exchangeability\_needed} (\texttt{WitCert/Conformal.lean:220}); \texttt{Conformal.le\_thr\_of\_not\_strictMax} (\texttt{WitCert/Conformal.lean:113}); \texttt{Conformal.mem\_strictMax} (\texttt{WitCert/Conformal.lean:69}); \texttt{Conformal.prob\_le\_one} (\texttt{WitCert/Conformal.lean:51}); \texttt{Conformal.prob\_mono} (\texttt{WitCert/Conformal.lean:48}); \texttt{Conformal.prob\_nonneg} (\texttt{WitCert/Conformal.lean:45}); \texttt{Conformal.prob\_union\_le} (\texttt{WitCert/Conformal.lean:56}); \texttt{Conformal.strictMax\_pairwiseDisjoint} (\texttt{WitCert/Conformal.lean:75}); \texttt{Conformal.sum\_prob\_strictMax\_le\_one} (\texttt{WitCert/Conformal.lean:84}); \texttt{Conformal.ties\_break\_disjointness} (\texttt{WitCert/Conformal.lean:213}); \texttt{Conformal.total\_risk\_over\_random\_history} (\texttt{WitCert/Conformal.lean:169}); \texttt{CumLoss.admit\_implies\_realized\_within} (\texttt{WitCert/CumLoss.lean:210}); \texttt{CumLoss.budget\_floor} (\texttt{WitCert/CumLoss.lean:234}); \texttt{CumLoss.cumloss\_admission} (\texttt{WitCert/CumLoss.lean:132}); \texttt{CumLoss.gating\_conservative} (\texttt{WitCert/CumLoss.lean:195}); \texttt{CumLoss.hitProbP\_eq\_hitProb} (\texttt{WitCert/CumLoss.lean:86}); \texttt{CumLoss.hitProbP\_le\_one} (\texttt{WitCert/CumLoss.lean:39}); \texttt{CumLoss.hitProbP\_mono} (\texttt{WitCert/CumLoss.lean:57}); \texttt{CumLoss.nonconservative\_admit\_can\_exceed} (\texttt{WitCert/CumLoss.lean:223}); \texttt{CumLoss.prodProcess\_exp} (\texttt{WitCert/CumLoss.lean:111}); \texttt{CumLoss.tailTerm\_mono} (\texttt{WitCert/CumLoss.lean:184}); \texttt{Ledger.telescope\_budget\_le} (\texttt{WitCert/Ledger.lean:73}); \texttt{Ledger.telescope\_sum\_le\_one} (\texttt{WitCert/Ledger.lean:51}); \texttt{Ledger.w\_nonneg} (\texttt{WitCert/Ledger.lean:34}); \texttt{McDiarmid.BddDiffAt.c\_nonneg} (\texttt{WitCert/McDiarmid.lean:72}); \texttt{McDiarmid.addF\_nonneg} (\texttt{WitCert/McDiarmid.lean:443}); \texttt{McDiarmid.band\_mean\_le} (\texttt{WitCert/McDiarmid.lean:515}); \texttt{McDiarmid.condE\_add} (\texttt{WitCert/McDiarmid.lean:387}); \texttt{McDiarmid.condE\_addF} (\texttt{WitCert/McDiarmid.lean:452}); \texttt{McDiarmid.condE\_bdd\_diff} (\texttt{WitCert/McDiarmid.lean:79}); \texttt{McDiarmid.condE\_congr} (\texttt{WitCert/McDiarmid.lean:408}); \texttt{McDiarmid.condE\_const\_mul} (\texttt{WitCert/McDiarmid.lean:51}); \texttt{McDiarmid.condE\_exp\_le} (\texttt{WitCert/McDiarmid.lean:201}); \texttt{McDiarmid.condE\_finset\_sum} (\texttt{WitCert/McDiarmid.lean:414}); \texttt{McDiarmid.condE\_mono} (\texttt{WitCert/McDiarmid.lean:41}); \texttt{McDiarmid.condE\_nonneg} (\texttt{WitCert/McDiarmid.lean:399}); \texttt{McDiarmid.condE\_sq\_le} (\texttt{WitCert/McDiarmid.lean:495}); \texttt{McDiarmid.eprocess\_factor\_mean\_le\_one} (\texttt{WitCert/McDiarmid.lean:556}); \texttt{McDiarmid.mcdiarmid} (\texttt{WitCert/McDiarmid.lean:310}); \texttt{McDiarmid.mcdiarmid\_radius} (\texttt{WitCert/McDiarmid.lean:359}); \texttt{McDiarmid.sum\_exp\_le\_of\_mean\_zero} (\texttt{WitCert/McDiarmid.lean:107}); \texttt{McDiarmid.witness\_mean\_le} (\texttt{WitCert/McDiarmid.lean:542}); \texttt{QuantityKind.sample\_mean\_below\_threshold\_not\_certifying} (\texttt{WitCert/QuantityKind.lean:96}); \texttt{QuantityKind.scope\_mismatch\_blocks\_pass} (\texttt{WitCert/QuantityKind.lean:102}); \texttt{QuantityKind.totalAlpha\_nil} (\texttt{WitCert/QuantityKind.lean:113}); \texttt{QuantityKind.undetermined\_consistent\_with\_both} (\texttt{WitCert/QuantityKind.lean:86}); \texttt{QuantityKind.undetermined\_costs\_nothing} (\texttt{WitCert/QuantityKind.lean:117}); \texttt{Radius.det\_shift\_sound} (\texttt{WitCert/Radius.lean:226}); \texttt{Radius.g\_pos} (\texttt{WitCert/Radius.lean:47}); \texttt{Radius.g\_zero} (\texttt{WitCert/Radius.lean:54}); \texttt{Radius.hasDerivAt\_g} (\texttt{WitCert/Radius.lean:56}); \texttt{Radius.hasDerivAt\_gd} (\texttt{WitCert/Radius.lean:66}); \texttt{Radius.hasDerivAt\_phi} (\texttt{WitCert/Radius.lean:81}); \texttt{Radius.log\_two\_point\_mgf\_le} (\texttt{WitCert/Radius.lean:145}); \texttt{Radius.phi\_deriv\_le\_quarter} (\texttt{WitCert/Radius.lean:100}); \texttt{Radius.phi\_deriv\_nonneg} (\texttt{WitCert/Radius.lean:94}); \texttt{Radius.phi\_le\_quarter\_mul} (\texttt{WitCert/Radius.lean:111}); \texttt{Radius.phi\_zero} (\texttt{WitCert/Radius.lean:79}); \texttt{Radius.quarter\_mul\_le\_phi} (\texttt{WitCert/Radius.lean:128}); \texttt{Radius.second\_deriv\_num\_eq} (\texttt{WitCert/Radius.lean:89}); \texttt{Radius.sq\_sum\_weighted\_le} (\texttt{WitCert/Radius.lean:212}); \texttt{Radius.sr\_mgf\_le} (\texttt{WitCert/Radius.lean:190}); \texttt{Radius.two\_point\_mgf\_le} (\texttt{WitCert/Radius.lean:184}); \texttt{Radius.two\_point\_variance} (\texttt{WitCert/Radius.lean:204}); \texttt{SharedRead.probE\_mono\_event} (\texttt{WitCert/SharedRead.lean:59}); \texttt{SharedRead.probE\_union\_le} (\texttt{WitCert/SharedRead.lean:48}); \texttt{SharedRead.shared\_read\_sound} (\texttt{WitCert/SharedRead.lean:79}); \texttt{Ville.hitProb\_nonneg} (\texttt{WitCert/Ville.lean:45}); \texttt{Ville.prodProcess\_supermartingale} (\texttt{WitCert/Ville.lean:106}); \texttt{crossSum\_comm} (\texttt{WitCert/Bridges.lean:716}); \texttt{crossSum\_indep\_meanzero\_eq\_zero} (\texttt{WitCert/Bridges.lean:784}); \texttt{crossSum\_sum\_left} (\texttt{WitCert/Bridges.lean:723}); \texttt{cumulative\_output\_tv} (\texttt{WitCert/Bridges.lean:418}); \texttt{exp\_sub\_one\_le\_mul\_exp} (\texttt{WitCert/Bridges.lean:129}); \texttt{gate\_threshold\_for\_sla} (\texttt{WitCert/Bridges.lean:206}); \texttt{linear\_propagation\_frobenius} (\texttt{WitCert/Bridges.lean:844}); \texttt{log\_jensen\_weighted} (\texttt{WitCert/Bridges.lean:904}); \texttt{massweighted\_prodenv\_bound} (\texttt{WitCert/Bridges.lean:950}); \texttt{massweighted\_topk\_bound} (\texttt{WitCert/Bridges.lean:920}); \texttt{one\_sub\_BC\_eq} (\texttt{WitCert/Bridges.lean:505}); \texttt{pool\_linf} (\texttt{WitCert/Contract.lean:189}); \texttt{propagated\_second\_moment\_le} (\texttt{WitCert/Bridges.lean:853}); \texttt{rank\_prefix\_stable} (\texttt{WitCert/Contract.lean:208}); \texttt{residual\_second\_moment\_indep} (\texttt{WitCert/Bridges.lean:821}); \texttt{residual\_second\_moment\_le} (\texttt{WitCert/Bridges.lean:758}); \texttt{residual\_second\_moment\_orthogonal} (\texttt{WitCert/Bridges.lean:737}); \texttt{score\_perturbation\_l2\_le} (\texttt{WitCert/Bridges.lean:866}); \texttt{sentinel\_rounds} (\texttt{WitCert/Contract.lean:225}); \texttt{served\_tv\_le\_of\_gate} (\texttt{WitCert/Bridges.lean:188}); \texttt{served\_tv\_le\_of\_gate\_tanh} (\texttt{WitCert/Bridges.lean:390}); \texttt{served\_tv\_le\_subgaussian} (\texttt{WitCert/Bridges.lean:634}); \texttt{served\_tv\_mean\_le\_massweighted} (\texttt{WitCert/Bridges.lean:977}); \texttt{served\_tv\_mean\_le\_omega\_subgaussian} (\texttt{WitCert/Bridges.lean:1049}); \texttt{softmax\_nonneg} (\texttt{WitCert/Bridges.lean:67}); \texttt{softmax\_sum} (\texttt{WitCert/Bridges.lean:73}); \texttt{softmax\_tilt} (\texttt{WitCert/Bridges.lean:79}); \texttt{softmax\_tv\_bridge\_tanh} (\texttt{WitCert/Bridges.lean:362}); \texttt{sqrt\_jensen\_weighted} (\texttt{WitCert/Bridges.lean:886}); \texttt{sum\_exp\_pos} (\texttt{WitCert/Bridges.lean:70}); \texttt{tv\_affine\_relax} (\texttt{WitCert/Bridges.lean:144}); \texttt{tv\_chord\_bound} (\texttt{WitCert/Bridges.lean:235}); \texttt{tv\_le\_hellinger} (\texttt{WitCert/Bridges.lean:445}); \texttt{tv\_le\_subgaussian} (\texttt{WitCert/Bridges.lean:548}); \texttt{tv\_le\_tanh} (\texttt{WitCert/Bridges.lean:271}); \texttt{var\_p\_delta\_le\_trace\_cov} (\texttt{WitCert/Bridges.lean:666}).\par}

\end{document}